\documentclass[preprints,article,accept,moreauthors]{Definitions/mdpi} 
\firstpage{1} 
\pubvolume{1}
\issuenum{1}
\articlenumber{0}
\pubyear{2026}
\copyrightyear{2026}
\datereceived{ } 
\daterevised{ } % Comment out if no revised date
\dateaccepted{ } 
\datepublished{ } 
\newcommand{\modified}[1]{\textcolor{black}{#1}}

\Title{Boosting Automatic Exercise Evaluation Through Musculoskeletal Simulation-Based IMU Data Augmentation}

\Author{Andreas Spilz $^{1}$\orcidA{}, Heiko Oppel $^{1}$\orcidB{} and Michael Munz $^{1}$*\orcidC{}}

\AuthorNames{Andreas Spilz, Heiko Oppel and Michael Munz}

\address[1]{%
$^{1}$ \quad AI for Sensor Data Analytics Research Group, Ulm University of Applied Sciences, Ulm, 89081, Germany}

\corres{Correspondence: Michael.Munz@thu.de}

\abstract{Automated evaluation of movement quality can enhance physiotherapeutic treatment and sports training by providing objective, real-time feedback. However, deep learning models that assess movements captured by inertial measurement units (IMUs) are often limited by data scarcity, class imbalance, and label ambiguity. We present a data augmentation method that generates IMU data using musculoskeletal simulations integrated with systematic modifications of movement trajectories. The approach enforces anatomically plausible kinematic constraints and enables automatic labeling by combining inverse kinematic parameters with a knowledge-based evaluation strategy. Across four datasets of varying complexity, augmented variants closely resemble real-world data and contribute to gains in classification accuracy, generalization to unseen subjects, and patient-specific fine-tuning from few examples. The magnitude of these gains varies with dataset properties, in particular class balance and label ambiguity. These findings indicate that musculoskeletal simulation-based augmentation can address common challenges faced by deep learning applications in physiotherapeutic exercise evaluation.}


\addhighlights{yes}
\renewcommand{\addhighlights}{%

\noindent\textbf{What are the main findings?}
 \begin{itemize}[labelsep=2.5mm,topsep=-3pt]
 \item A musculoskeletal simulation-based method generates anatomically plausible IMU data from IMU recordings alone and labels the augmented variants automatically by combining inverse kinematic parameters with a knowledge-based evaluation strategy, without motion capture, video, or manual annotation.
 \item Adding augmented data to real training data improves classification accuracy and generalization to unseen subjects, with the largest gains under leave-one-subject-out evaluation, including subjects whose training data lack an entire movement class.
 \end{itemize}\vspace{3pt}
\textbf{What are the implications of the main findings?}
 \begin{itemize}[labelsep=2.5mm,topsep=-3pt]
 \item The method reduces the dependence on large annotated IMU datasets and mitigates class imbalance in automated movement quality assessment.
 \item Augmented data enable patient-specific fine-tuning from only a few real repetitions, supporting individualized assessment models in physiotherapy and training.
 \end{itemize}
}

\begin{document}

%%%%%%%%%%%%%%%%%%%%%%%%%%%%%%%%%%%%%%%%%%

\section{Introduction}
Home-based training represents a vital component of many physiotherapeutic treatment strategies and has been shown to enhance patient outcomes~\citep{ashari_effectiveness_2016, latham_effect_2014, gelaw_effectiveness_2020, flynn_home-based_2019}. However, these exercises are typically performed without professional oversight, presenting several challenges. The lack of supervised guidance can negatively impact adherence~\citep{argent_patient_2018} to the prescribed regimen and lead to incorrect execution of the prescribed exercises~\citep{faber_majority_2015}. Consequently, progress may be slowed, and in some cases, improper execution can lead to inappropriate loading or even injury. These issues can be effectively addressed by systems that monitor and analyze movements, providing real-time feedback to the user. Such systems have demonstrated the ability to improve adherence~\citep{lang_digital_2022} and hold promise for enhancing the safety and efficacy of unsupervised physiotherapeutic exercises.

To address this challenge, in an earlier work we developed a sensor system that integrates inertial measurement units (IMUs) with deep learning algorithms to enable automated evaluation of movement patterns~\citep{spilz_automatic_2023}. IMUs offer a portable and cost-effective means of capturing three-dimensional motion data in real time, while deep learning methods excel at identifying subtle patterns and complex relationships within large datasets. This combined approach was tested using exercises from Functional Movement Screening (FMS), a widely recognized protocol for evaluating key aspects of movement quality, including stability and mobility~\citep{cook_functional_2014, cook_functional_2014-1}. We showed that convolutional-recurrent networks can learn exercise-specific spatiotemporal patterns from IMU data. However, performance declined substantially for unseen participants in leave-one-subject-out (LOSO) evaluations, indicating that limited data diversity together with class imbalance constrains generalization.

The main reason for the observed performance decline is the limited availability of training data, a constraint shared across the field. Datasets collected by our group~\citep{spilz_automatic_2023} and others~\citep{xing_functional_2022, wu_development_2020} exhibit three key issues: a small quantity of data points per exercise, a pronounced class imbalance, and a limited number of recorded subjects, resulting in limited diversity in movement patterns. These data limitations impede the development of robust machine learning models for motion assessment and often lead to poor generalization, especially on unseen subjects~\citep{spilz_automatic_2023, scheurer_comparing_2020, kianifar_automated_2017}.

Comprehensive data collection efforts alone cannot resolve this issue. In addition to the substantial monetary and temporal resources required, it is highly challenging to recruit appropriate test participants representative of the full spectrum of movement performance. Acquiring samples of perfectly executed repetitions requires test subjects with extraordinary abilities, which are scarce within the general population. In contrast, intentionally inducing severely flawed repetitions carries an ethically unacceptable risk of injury. 

As an alternative, a growing body of work generates synthetic or augmented IMU data rather than collecting more. These approaches differ in two respects that are decisive for exercise evaluation: the input data they require, and whether they can assign a class label to a newly generated movement. The following review is ordered along these two axes.

Some of these generate IMU data based on data from other domains such as motion capture (MoCap) data~\citep{sharifi_renani_use_2021, mundt_estimation_2020, uhlenberg_synhar_2024}, video data~\citep{kwon_imutube_2020} or pose data~\citep{zolfaghari_sensor_2024}. These approaches are promising, yet they only work if the corresponding MoCap, video, or pose data are available. Such data, however, is often scarce or unavailable, particularly in the context of specialized physiotherapeutic exercises.

In contrast to these methods, purely data-driven approaches use neural networks to generate synthetic data, for example by using GANs~\citep{norgaard_synthetic_2018, mohammadzadeh_cgan-based_2025, zhao_improving_2022} or text-to-IMU applications such as IMU-GPT~\citep{leng_generating_2023, leng_imugpt_2024}. However, these models likewise depend on having enough examples in their training dataset, particularly for minority classes, to learn a valid representation. If a specific class is undersampled, these purely data-driven techniques may fail to capture its essential characteristics. Moreover, most of these approaches \modified{lack mechanisms to keep generated movements within the anatomically valid ranges of joint motion}~\citep{norgaard_synthetic_2018, mohammadzadeh_cgan-based_2025, zhao_improving_2022, leng_generating_2023}, and none of them ensure correct class assignment for the generated samples~\citep{norgaard_synthetic_2018, mohammadzadeh_cgan-based_2025, zhao_improving_2022, leng_generating_2023, leng_imugpt_2024}. Such limitations are especially problematic when augmenting an underrepresented class with a large volume of synthetic data: if the generated examples are biomechanically invalid or incorrectly classified, the model could learn unrealistic patterns that are not representative of real-world scenarios. 

A further line of work leverages musculoskeletal models or physics-based simulations to ensure the biomechanical plausibility of generated examples. These simulation-based methods split into two families according to how they treat class labels. Many of these approaches inherently preserve the original class labels (``label-preserving''). Rather than synthesizing data examples associated with different labels, their primary aim is to increase the variability of training samples within each existing class through different mechanisms. Uhlenberg et al.~\citep{uhlenberg_synhar_2024} and Mundt et al.~\citep{mundt_estimation_2020} generate additional samples per class by varying the position and orientation of the simulated IMUs relative to the body, leaving the underlying movement unchanged. Chandrasekaran et al.~\citep{chandrasekaran_gait_2023} scale the anthropometric proportions of the musculoskeletal model to represent diverse body morphologies, again without altering the movement itself. Tang et al.~\citep{tang_synthetic_2024} combine sensor pose perturbation and model scaling for fall detection, using multi-camera video instead of motion capture as the source of body movement information. Closest to our approach, Oishi et al.~\citep{oishi_physically_2026} apply magnitude and time scaling to joint rotation trajectories to introduce physically plausible variations in movement amplitude and speed, complemented by sensor-placement and hardware-related variations, for human activity recognition. Their parameter ranges are chosen conservatively, such that augmented samples remain within the original activity class. These methods further differ in the input modality required during data generation. Several rely on motion capture~\citep{mundt_estimation_2020, uhlenberg_synhar_2024} or multi-camera video recordings~\citep{tang_synthetic_2024}, while Oishi et al.~\citep{oishi_physically_2026} require paired IMU and motion data for an initial parameter identification stage. Our approach, by contrast, operates exclusively on IMU recordings.

A second family of approaches explicitly produces non-label-preserving augmented data. Dorschky et al.~\citep{dorschky_cnn-based_2020} and Renani et al.~\citep{sharifi_renani_use_2021} synthesize IMU data together with the corresponding biomechanical target variables, such as joint angles, joint moments, or ground reaction forces, which are produced as direct outputs of the musculoskeletal simulation and therefore become available alongside the synthesized IMU data without any further processing.
In automated physiotherapeutic exercise evaluation, the relationship between simulation outputs and class labels is less direct. Each class is defined by a combination of kinematic criteria that must be evaluated jointly. The class label of an augmented repetition therefore cannot be read off a single simulation output but requires an additional step in which several derived parameters are extracted and combined according to the criteria that define each class.

Building upon the existing research, we propose a novel approach tailored to augment IMU datasets within the domain of automatic exercise evaluation, explicitly addressing the aforementioned limitations regarding data availability, biomechanical plausibility, and label assignment complexity. In summary, this paper focuses on the following key aspects:

\begin{itemize}
    \item \textbf{Novel data augmentation method}

We introduce a novel data augmentation method to \modified{generate IMU data} by integrating musculoskeletal simulations with targeted modifications of movement trajectories. This approach intentionally generates examples across multiple classes while \modified{respecting physiological joint-range limits.} Notably, the method operates exclusively on IMU data and does not depend on motion capture, video, or pose data.
    \item \textbf{Automatic labeling approach}

Our method combines inverse kinematics parameters with a knowledge-based evaluation strategy to derive labels automatically, removing the need for manual annotation of augmented repetitions.
    \item \textbf{Extensive evaluation of the characteristics of augmented data and their influence on the performance of an automated evaluation classifier}

We rigorously evaluate both the characteristics and practical utility of the generated data. Employing dimensionality reduction, we verify the realism and variability of the augmented dataset compared to real-world data. A dedicated process-validation experiment rules out pipeline-induced distribution artifacts. A robustness analysis assesses the stability of the results across varying augmentation configurations. Controlled experiments across five evaluation paradigms assess how incorporating augmented data influences the accuracy and generalization performance of a neural network for automatic exercise evaluation.

    \item \textbf{Cross-subject generalization and patient-specific fine-tuning}

We assess generalization to unseen subjects through leave-one-subject-out cross-validation. Building on this, we examine the effectiveness of augmented data for patient-specific fine-tuning scenarios of pretrained models. Our experiments quantify how augmented samples enhance personalized predictions, particularly in scenarios with limited subject-specific training examples.
\end{itemize}

%%%%%%%%%%%%%%%%%%%%%%%%%%%%%%%%%%%%%%%%%%
\section{Materials and Methods}

In this section, we first provide a detailed description of the datasets used in our study, highlighting their properties, challenges, and the respective measurement procedures. Subsequently, we outline our novel augmentation method, including preprocessing of IMU data, systematic modification of movement orientations, inverse kinematics-based validation, and the automatic labeling approach. Finally, we present our comprehensive evaluation strategy, comprising data characterization, the neural network architecture, experimental setups assessing the utility of augmented data, and a targeted exploration of patient-specific fine-tuning scenarios.

\subsection{Used datasets}
In this study, we evaluate our novel data augmentation approach using four datasets collected by our team through two separate measurement studies~\citep{spilz_automatic_2023, spilz_gaitex_2025}. For more details about the measurement process and study design, please refer to the original publications. These datasets comprise four distinct exercises: two FMS tasks, the Deep Squat (DS) and the Hurdle Step (HS), and two rehabilitation exercises commonly used in the treatment of foot drop, the Resisted Dorsiflexion (RD) and the Resisted Gait Simulation (RGS).

\subsubsection{Dataset properties and challenges}

The selected exercises exhibit variability in difficulty for neural network analysis based on two primary properties: class distribution and ambiguity in labeling.

\begin{figure}[H]
\centering
\includegraphics[width=\linewidth]{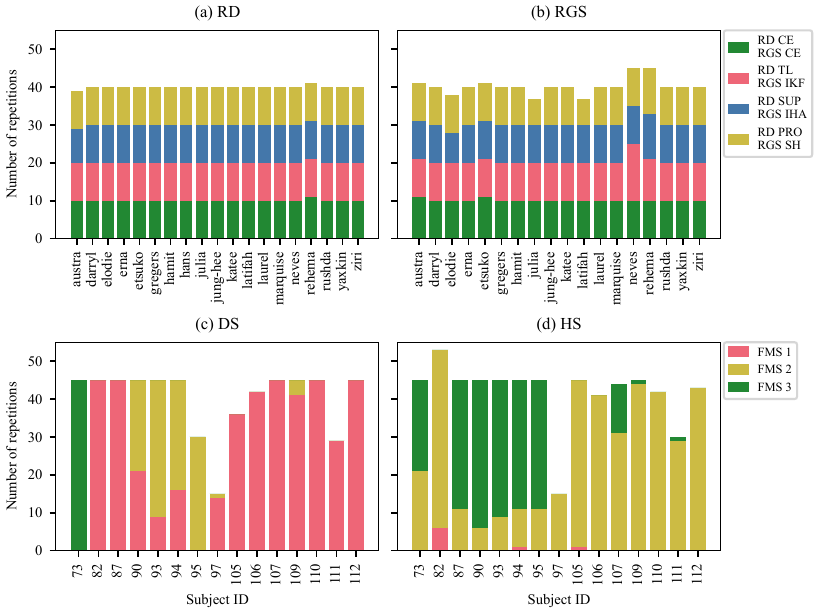}
\caption{Class distribution as number of repetitions across subjects for the RD, RGS, DS, and HS exercises.\\
(a) RD with the execution variants CE, TL, SUP, and PRO. (b) RGS with CE, IKF, IHA, and SH. (c) DS and (d) HS, both with their respective FMS classes. Subject labels are pseudonyms assigned during data acquisition.}
\label{fig:rep_per_subject_and_label}
\end{figure}

Class distribution shapes complexity at both the overall and the individual-participant level. The RD and RGS datasets represent favorable cases, each with a balanced distribution across four categories. For RD these are correct execution (CE), toe lifted (TL), supination (SUP), and pronation (PRO), and for RGS correct execution (CE), increased knee flexion (IKF), increased hip abduction (IHA), and static hip (SH). RD comprises approximately 760 repetitions across 19 participants and RGS approximately 720 across 18, both showing comparable per-class counts overall and at the subject level~\citep{spilz_gaitex_2025} (see Figure~\ref{fig:rep_per_subject_and_label}~a,b).

The DS and HS datasets present more complex scenarios. Both use the three FMS classes, which rate a repetition as showing severe movement deficiencies (1), moderate execution with minor compensations (2), or optimal execution (3). The overall distribution is imbalanced, with most participants contributing repetitions from only one or two classes. DS comprises roughly 600 repetitions and HS roughly 620, each from 15 participants (see Figure~\ref{fig:rep_per_subject_and_label}~c,d). The imbalance is stronger for HS, making it the more demanding of the two for model training and evaluation.

Label assignment forms the second property affecting complexity. For RD and RGS, labels were assigned by instruction: participants were taught each movement variation in advance and replicated it on command, with the command defining the label. This procedure yields reliable labels in most cases, though individual repetitions may deviate from the prescribed variation.

\begin{figure}[H]
\centering
\includegraphics[width=\linewidth]{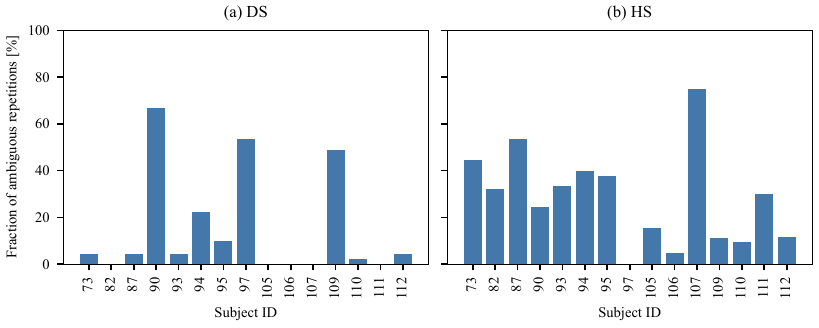}
\caption{Ambiguous label assignments for the DS and HS exercises.\\
Illustrated are the fractions of exercise repetitions that were deemed ambiguous (i.e., repetitions with differing ratings among independent physiotherapists) for individual participants. (a) DS exercise. (b) HS exercise.}
\label{fig:ambivalent_percentage}
\end{figure}

For DS and HS, labels relied on assessments by three independent physiotherapists following the FMS criteria of Cook et al.~\citep{cook_functional_2014, cook_functional_2014-1}. The subjective interpretation of these guidelines introduces ambiguity, and the fraction of ambiguous repetitions, those with disagreement among raters, varies between the two exercises (see Figure~\ref{fig:ambivalent_percentage}). DS shows a notable level of ambiguity, while HS reaches the highest degree among all four datasets. Prior work attributes this to the inherent subjectivity of the FMS criteria~\citep{shultz_test-retest_2013}. Overall, dataset complexity increases from RD and RGS through DS to HS, driven by class imbalance and label ambiguity.

\subsubsection{Measurement procedure}

\modified{The FMS datasets (DS and HS) were recorded with 15 Shimmer3 IMUs (ShimmerSensing, Dublin, Ireland). Sensor count and placement followed the established Xsens MVN full-body configuration for IMU motion capture~\citep{roetenberg}. The RD and RGS datasets were recorded with nine Xsens MVN Awinda IMUs (Movella, El Segundo, California, USA). For both exercises, the full-body scheme was reduced to the lower limbs and pelvis, and an additional toe-mounted IMU was placed on the right foot to better resolve foot kinematics. Both systems include tri-axial accelerometers, gyroscopes, and magnetometers. Shimmer3 sensors sampled at 120 Hz with ranges of $\pm16$~g (accelerometer), $\pm200^{\circ}$/s (gyroscope), and $\pm49.5$~Ga (magnetometer). Xsens sensors sampled at 100~Hz with ranges of $\pm16$~g, $\pm2000^{\circ}$/s, and $\pm1.9$~Ga, respectively. Further details on sensor placement and the measurement procedures are provided in the accompanying publications~\citep{spilz_automatic_2023, spilz_gaitex_2025}.}

\subsubsection{Ethical approval}
All IMU recordings used in this study were obtained from two previously published measurement studies~\citep{spilz_automatic_2023, spilz_gaitex_2025} collected under ethically approved research protocols. Ethical approval was granted by the Ethics Committee of Ulm University of Applied Sciences (reference numbers 2021-01 and 2024-01). Written informed consent was obtained from all participants prior to participation in the respective data collections. This includes consent for data acquisition, storage, and anonymized analysis for research purposes.

\subsection{Augmentation process}
Our augmentation process enables the generation of modified versions of a given movement exercise based on IMU data. These variations can either represent the same movement quality label (e.g., the same FMS score) or be altered in a way that results in a different label. Since we train our neural networks on orientations rather than raw IMU data, both the input and output of our augmentation process are represented as quaternions. The following sections provide a detailed description of each step involved in the augmentation process. 

The code used to implement the described augmentation steps is available at: \newline
\url{https://github.com/ai-for-sensor-data-analytics-ulm/imu_augment_sim}.

\subsubsection{Necessary preprocessing steps}
To enable the augmentation process, several preprocessing steps are required. First, the orientation of each IMU must be estimated from the raw sensor data. Depending on the dataset, different algorithms were applied. For the FMS exercises, orientation estimation was performed using the Madgwick~\citep{madgwick_estimation_2011} orientation filter ($\beta=0.033$), which utilized accelerometer and gyroscope data to provide a stable orientation estimate~\citep{spilz_automatic_2023}. In contrast, the RD and RGS datasets relied on the proprietary XKF3hm algorithm, which is implemented on the MVN Awinda sensors from Movella~\citep{paulich_xsens_2018} and computes orientation directly on the device.

The estimated quaternions $T_{IMU,s}$ represent the orientation of each IMU in lab space and, with it, the orientation of the attached body segment $s$. Variations in sensor placement across subjects make these raw orientations inconsistent. To remove this effect, we define a unified coordinate system per body segment and compute, at the start of each recording, an initial transformation $T_{offset,s}$ that maps the IMU orientation onto its segment coordinate system. The segment orientation $T_s$ then follows as:

$$T_s = T_{offset,s} \cdot T_{IMU,s}$$

The exact computation of $T_{offset,s}$ is detailed in prior research~\citep{spilz_automatic_2023, spilz_gaitex_2025}, and other sensor-to-segment calibration methods are equally applicable.

\subsubsection{Targeted modification of orientations}
Following the preprocessing steps, the segment orientation trajectories $T_s$ are systematically modified to introduce variations in movement patterns. To achieve this, the quaternion-based orientation data are first converted into Euler angles. The time-dependent progression of each Euler angle of segment $s$ is defined as:

$$e_s[t]=
\begin{bmatrix}
\phi[t] \\
\theta[t] \\
\psi[t]
\end{bmatrix},
$$
corresponding to rotations around the respective axes (roll, pitch, yaw).
To generate augmented motion trajectories, we modify the Euler angles using two augmentation parameters: an offset vector $\beta$, defining the initial posture, and a scaling vector $\alpha$, which adjusts the relative range of motion to match the desired target range $\delta$. The augmented trajectory is computed as follows:
$$e_{s,aug}[t] = (e_s[t] - e_s[0]) \cdot \alpha + \beta,
$$
with 
$$
\alpha = 
\begin{bmatrix}
\frac{\delta_\phi}{\max(e_\phi[t])- \min(e_\phi[t])} \\
\frac{\delta_\theta}{\max(e_\theta[t])- \min(e_\theta[t])} \\
\frac{\delta_\psi}{\max(e_\psi[t])- \min(e_\psi[t])} \\

\end{bmatrix}, \quad 
\beta= 
\begin{bmatrix}
\beta_\phi \\
\beta_\theta \\
\beta_\psi
\end{bmatrix}.
$$

The augmentation parameters $\beta$ and $\delta$ are drawn from two distinct multivariate normal distributions, which are furthermore specific to a given rating, exercise, and segment. 

These distributions are constructed by calculating mean and standard deviation of offsets and ranges for each Euler angle component $c$ from available repetitions:
$$\beta_c = e_c[0],   \quad  \delta_c = max(e_c[t]) - min(e_c[t]).$$

Once these distributions have been established, they can be used to generate augmented Euler angle trajectories. If the augmentation parameters are sampled from the distributions corresponding to the original rating, the augmentation introduces moderate variations while preserving the overall movement characteristics. In contrast, selecting augmentation parameters from a distribution associated with a different rating allows for more significant modifications, enabling the generation of movement variations that reflect different execution qualities.

The parameters of these normal distributions can be established using two approaches: (1) empirical estimation from recorded datasets as shown above and (2) expert knowledge from experienced physiotherapists. Ideally, both approaches are combined: statistical patterns extracted from recorded data provide a quantitative foundation, while expert knowledge ensures that the distributions align with realistic movements. This alignment is particularly crucial in cases where the sample size for a specific class is limited. In this study, the distributions were initially generated based on available data and subsequently reviewed by experts for validation.

\begin{figure}[H]
\centering
\includegraphics[width=\linewidth]{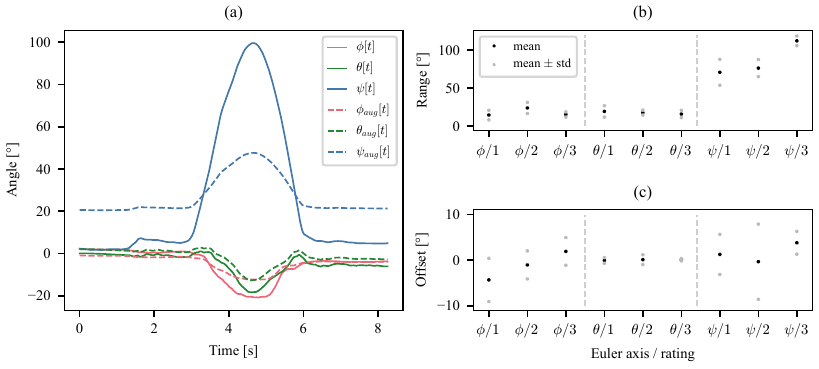}
\caption{Targeted modification of orientations, shown for one DS repetition and the Euler angle trajectory of the right femur.\\
(a) Original (solid) and augmented (dashed) trajectories, with augmentation parameters sampled from the corresponding multivariate distributions. (b) Range and (c) offset distribution parameters per Euler axis and FMS rating category.}
\label{fig:aug_process}
\end{figure}

Figure~\ref{fig:aug_process} presents a visual illustration of the proposed augmentation process. It depicts the baseline and augmented Euler angle trajectories for the femur segment, along with the associated distribution parameters used to generate the augmented motions.

Following the transformation, the modified Euler angles are converted back into quaternions. Using Euler angles for augmentation offers key advantages, as they provide an intuitive representation of movement that aligns well with physiotherapeutic expertise. This interpretability is essential for validating the generated variations.

The rationale behind this approach is that variations in physiotherapeutic exercise execution are often characterized by subtle deviations in the initial posture and range of motion. By adjusting the offset parameter, variations in initial posture are introduced, while modifications to the scaling factor alter the range of motion. As a result, this method enables the generation of diverse movement instances that may either remain within the same evaluation class or represent different classes.

\subsubsection{Inverse kinematics}
The transformation of the orientation data, as described above, only considers individual body segments and does not account for the kinematic dependencies between adjacent segments. To respect the \modified{anatomically valid joint constraints} of the human body and thereby generate more realistic augmented samples, we perform an inverse kinematics computation based on the generated orientation trajectories and a suitable skeletal model.

For this purpose, we use the IMU Inverse Kinematics Toolbox in the open-source software OpenSim 4.5~\citep{delp_opensim_2007}. The skeletal model used is a customized version of the model by Rajagopal et al.~\citep{rajagopal_full-body_2016} that was adapted to the specific requirements of our application. We added a head segment with a three-degree-of-freedom neck joint and extended the knee, elbow, and wrist joints with additional degrees of freedom, so that the orientation of every instrumented segment can be reproduced. We released the previously locked subtalar, metatarsophalangeal, and wrist joints and widened the joint-range limits at the shoulder, ankle, subtalar, metatarsophalangeal, and forearm joints to cover the range of motion observed during the exercises. The adapted model is released as part of the accompanying code. The inverse kinematics algorithm in OpenSim calculates for each time step the pose of the model that best fits the given segment orientations, while adhering to the predefined \modified{joint-range} constraints of the model.

Because the modified orientation trajectories were already expressed in their respective segment coordinate systems, they directly drive the individual segments during the computation.

\begin{figure}[H]
\centering
\includegraphics[width=0.667\linewidth]{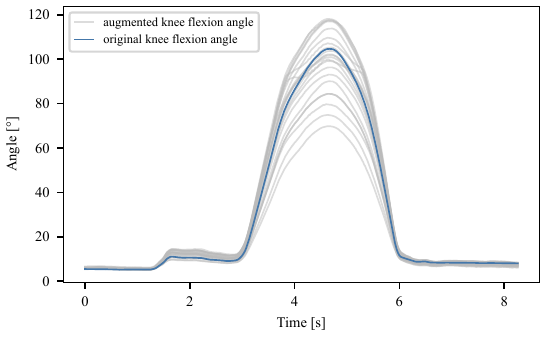}
\caption{Visualization of knee flexion angles for one original and 20 augmented variants of it.}
\label{fig:results_aug_process}
\end{figure}

To illustrate the resulting variability of the generated movements, Figure~\ref{fig:results_aug_process} shows the knee flexion angle of the right leg for one original repetition and 20 augmented variants produced by the proposed augmentation pipeline.

\subsubsection{Automatic labeling}
The proposed augmentation technique does not preserve the original class labels, so the class of each generated repetition must be determined after generation. The sampling distributions are class-specific, but they do not guarantee the intended label for two reasons. First, the inverse kinematics computation can alter segment orientations and shift a movement into a different category. Second, parameters sampled far from the class-specific mean can yield examples that align more closely with another class.

Manual relabeling of every generated movement by expert physiotherapists would be reliable but is too time-consuming and costly to be practical. We therefore developed an automated evaluation method that assigns the class from domain knowledge about the exercise and the kinematic information derived from the inverse kinematics computation.

\begin{figure}[H]
\centering
\includegraphics[width=\linewidth]{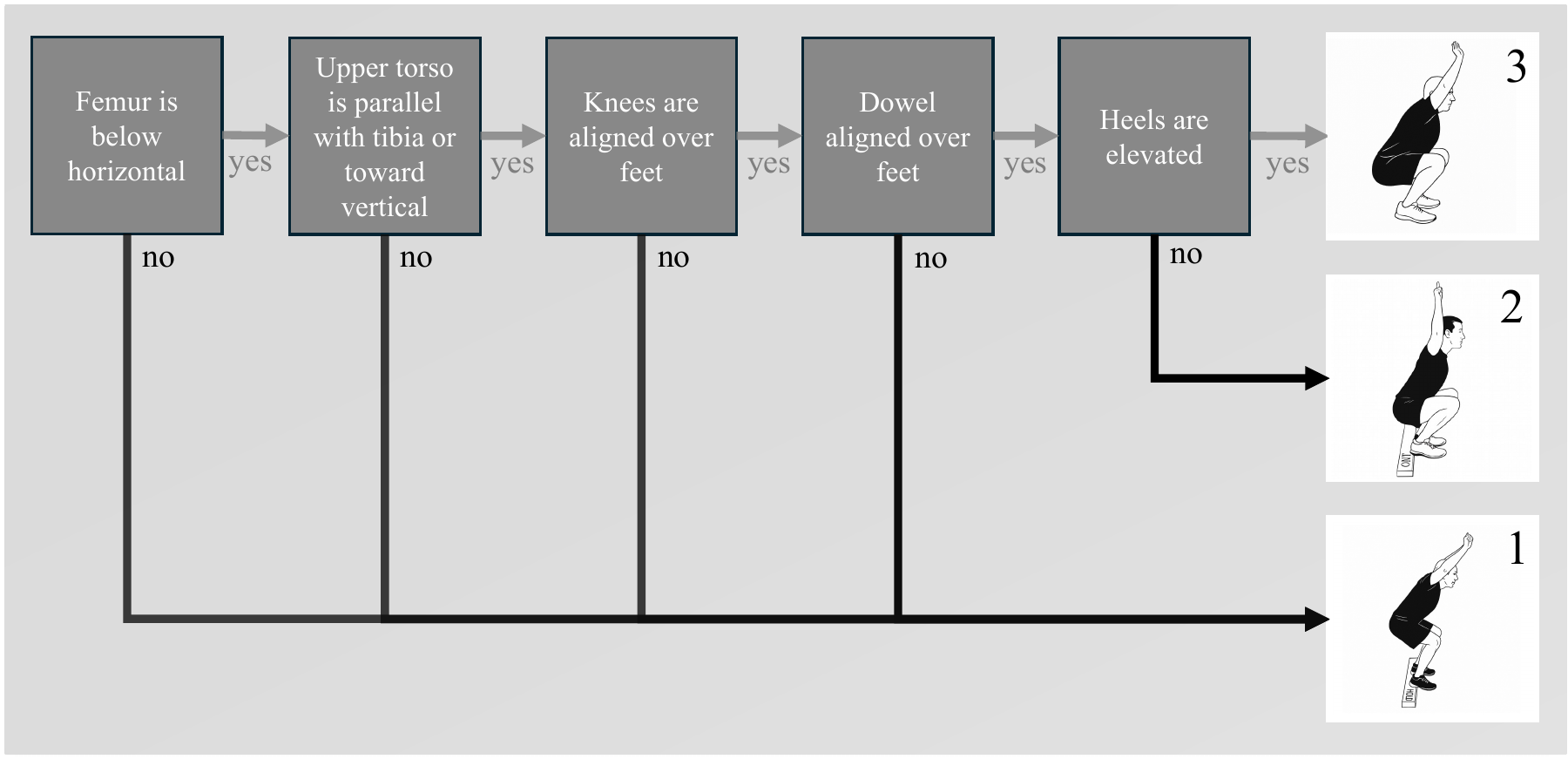}
\caption{Labeling process for the DS as defined by Cook et al.~\citep{cook_functional_2014, cook_functional_2014-1}.\\
Each exercise is characterized by multiple criteria (e.g., ``Femur is below horizontal'' or ``Heels are elevated''). Their fulfillment or violation determines the final exercise quality label.}
\label{fig:labeling_process_ds}
\end{figure}

This domain knowledge is encoded in the predefined evaluation process of each FMS exercise and in the error-pattern definitions of RD and RGS. Each exercise is described by a set of criteria that are either met or unmet, together with a rule that assigns a label from the fulfilled and unfulfilled criteria (see Figure~\ref{fig:labeling_process_ds}). Each criterion maps to a parameter computed from the musculoskeletal simulation, such as joint angles, segment positions, or distances between anatomical points.

During development, we found that certain evaluation criteria could not be translated unambiguously into fixed thresholds on the derived kinematic parameters. The DS criterion ``hands above feet'', for example, leaves the exact anatomical reference points and acceptable deviations open, which invites subjective interpretation. To obtain consistent labels for real and augmented data, we therefore optimize these thresholds against the available real data and their original physiotherapist-assigned labels.

The real data are passed through the same inverse kinematics computation as the augmented data and then through the automated classification framework. Agreement between the resulting automated labels and the physiotherapist-assigned labels is quantified by the geometric mean of the class-wise F1-scores over the $n$ classes, $GM_{F_1}$:

$$GM_{F_1} = \left( \prod_{i=1}^{n} F_{1,i} \right)^{\frac{1}{n}}.$$

We optimize the thresholds by random search: the ruleset remains fixed, while thresholds are sampled uniformly from the ranges spanned by the observed minimum and maximum parameter values in the real data. Across 10 million combinations per exercise, we select the set with the highest $GM_{F_1}$ for labeling the augmented data. The selected threshold sets were manually reviewed for plausibility and were considered appropriate for labeling the augmented repetitions. The resulting $GM_{F_1}$ scores for the four exercises are reported in Table~\ref{tab:1}.

\begin{table}[H]
\centering\caption{Best achieved geometric mean of F1-Scores $GM_{F_1}$ using the described random search methodology for the four used datasets.}
\label{tab:1}
\begin{tabular}{ccccc}
\toprule
Exercise & RD & RGS & DS & HS\\
\midrule
$GM_{F_1}$ & 0.88 & 0.83 & 0.84 & 0.75 \\
\bottomrule
\end{tabular}
\end{table}

\subsection{Evaluation}
This section describes the experiments conducted to characterize the augmented repetitions and investigate their impact on training a neural network.

\subsubsection{Generation of augmented examples}
Augmented examples were generated from the recorded repetitions for each of the possible classes, yielding approximately 12,000 augmented examples for the RD and RGS datasets and approximately 20,000 for the DS and HS datasets. The computation time required to generate one example corresponded roughly to the execution duration of the respective movement. This computational performance was determined on a desktop computer equipped with an AMD Ryzen 5 3600X 6-Core Processor operating at 3.79 GHz.
To ensure only unambiguous examples were created, a preliminary selection process was implemented. Only examples whose automatic labeling matched the class assigned by the selected distribution were retained for further analysis.

\subsubsection{Characteristics of real and augmented samples}
To gain an initial understanding of the characteristics of the augmented examples, we conducted a dimensionality reduction using the t-distributed Stochastic Neighbor Embedding (t-SNE) method~\citep{tsne}. This analysis was performed separately for each dataset, involving both the real data and a subset of 250 augmented examples per class. To ensure a balanced representation, the augmented examples were selected such that an equal number of examples per class were included for each subject.

The t-SNE analysis was carried out using the implementation provided by scikit-learn (version 1.8.0). Prior to applying t-SNE, the data were projected onto 50 principal components using Principal Component Analysis (PCA) to reduce computational cost and suppress noise. All t-SNE parameters were set to their default values. In preliminary runs, parameter variation was conducted within a sensible range to confirm that the observed characteristics were consistently detectable across different parameter configurations.

\subsubsection{Preprocessing}
All data samples (generated and real data) undergo the following preprocessing steps to be compatible with the neural network. The quaternion-based orientation sequences of the individual body segments are linearly interpolated to a standardized temporal length of 256 time steps, employing the Spherical Linear Interpolation technique (SLERP)~\citep{shoemake_animating_1985}. Subsequently, the interpolated orientation trajectories for each IMU are arranged in a row-wise format, constructing a tensor representation suitable for CNN processing.

\subsubsection{Neural network architecture and training details}
For our evaluation, we implemented a convolutional neural network (CNN) classifier to assess the effect of augmented data on model performance. The network receives as input a tensor of shape $(N_{\text{IMU}} \times 4) \times 256$, where $N_{\text{IMU}}$ denotes the number of IMUs, the factor of 4 corresponds to the quaternion components per IMU, and 256 is the number of time steps after SLERP interpolation.

The feature extraction stage consists of two consecutive convolutional blocks. The first block applies 32 filters with a kernel of $(4, 1)$ and stride $(4, 1)$. Since the four quaternion components of a single IMU jointly encode its orientation, this kernel aggregates them into a compact per-IMU representation before any temporal processing takes place. The second block applies 64 filters with a kernel of $(1, 8)$ and stride $(1, 6)$, capturing temporal patterns with a slight overlap along the time axis. Each convolutional layer is followed by batch normalization and a Rectified Linear Unit (ReLU) activation.

Following the convolutional stage, the feature maps are flattened and passed through a fully connected block consisting of a linear layer with 64 neurons, a ReLU activation, and a dropout layer with a rate of 0.2. The output layer contains as many neurons as there are classes in the respective dataset and applies a softmax activation to produce class probabilities. The resulting network is trained using cross-entropy loss.

The network was trained end-to-end using the following configuration. We employed the Adam optimizer with an initial learning rate of $10^{-4}$ and a weight decay of $10^{-3}$. Training was conducted with mini-batches of size 32 for a maximum of 2500 epochs using weighted cross-entropy loss, with per-class weights set inversely proportional to class frequency in the training set to account for class imbalance. A ReduceLROnPlateau scheduler was applied, halving the learning rate when no improvement in validation loss was observed for 10 consecutive epochs (minimum delta $10^{-3}$). An early stopping criterion terminated training if the validation loss did not improve for 30 consecutive epochs (minimum delta $10^{-3}$). The model checkpoint achieving the lowest validation loss was retained for final evaluation. The entire pipeline was implemented in PyTorch (version 2.9.1). These hyperparameters were not formally optimized but chosen empirically based on preliminary experiments, and kept identical across all datasets and evaluation paradigms.

\subsubsection{Experiments}

We conducted two sets of experiments. The first characterizes how augmented repetitions compare to real ones in terms of characteristics and variance, using four training--testing scenarios under 5-fold cross-validation. The second measures how combining real and augmented data affects generalization to unseen subjects, using two scenarios under leave-one-subject-out cross-validation (LOSOCV).

Across all scenarios that include augmented data, two augmented examples were added for each real example in both the training and validation sets, giving a 1:2 ratio of real to augmented samples. These examples were drawn uniformly across classes and subjects, and each was generated exclusively from the real examples of its own training partition, preventing information leakage between splits.

The first set comprises four scenarios, each named for the data used for training and testing:

\begin{itemize}
    \item TRTR (train real, test real): baseline for the classification task.
    \item TRATR (train real and augmented, test real): whether adding augmented samples to the real training set preserves or improves performance on real data.
    \item TATR (train augmented, test real): how well augmented data alone captures the variance of the real data.
    \item TRTA (train real, test augmented): whether the variance of the augmented data is already covered by the real training data.
\end{itemize}

These scenarios were evaluated with 5-fold cross-validation, with folds stratified so that each contained examples from all subjects and a comparable proportion of examples per class. In each iteration, four folds formed the training set, split 80/20 into training and validation, and the held-out fold was used exclusively for testing.

The second set uses LOSOCV, which reproduces the realistic case of a model applied to a subject absent from training. In each iteration, one subject was held out entirely for testing, while the remaining subjects' data were split 80/20 into training and validation. Depending on the subject count, this yields a 19-fold LOSOCV for RD, an 18-fold for RGS, and a 15-fold for DS and HS. We compared two scenarios:

\begin{itemize}
    \item TRTR-LOSO (train real, test real): baseline for generalization to an unseen subject.
    \item TRATR-LOSO (train real and augmented, test real): the effect of augmented data on that generalization.
\end{itemize}

\subsubsection{Control experiments}

Two control experiments examine the augmented data itself, separately from the main classification results.

The first is an augmentation pipeline validation. It tests whether the pipeline introduces artifacts that affect classifier performance independently of the intended movement variation. Real repetitions were passed through the complete pipeline, including preprocessing, inverse kinematics, and post-processing, but without any Euler angle modification ($\alpha = 1$, $\beta = 0$ for all segments and axes). These repetitions are processed but movement-preserving and keep their original labels. Using them in place of the augmented data in the TATR and TRTA setups defines two paradigms:

\begin{itemize}
    \item TPTR (train pipeline-processed, test real): whether pipeline-processed data alone reproduce performance on real data.
    \item TRTP (train real, test pipeline-processed): whether the real training data already cover the pipeline-processed data.
\end{itemize}

Since movement and labels are unchanged, a performance gap relative to the TRTR baseline would isolate a distribution shift caused by the pipeline alone.

The second is a robustness analysis. It measures how much augmented training data help the classifier cope with underrepresented execution variants. Both models are evaluated on augmented test data:

\begin{itemize}
    \item TRTA (train real, test augmented): a model trained on real data only.
    \item TRATA (train real and augmented, test augmented): a model trained on the combined real and augmented set.
\end{itemize}

The augmented examples broaden the coverage of plausible execution variants without leaving the anatomically valid range, so the comparison shows how well each model handles variations that are sparse in the real training data.

\subsubsection{Patient-specific fine-tuning}
Typically, the available datasets contain a limited number of participants, providing insufficient data to train neural networks capable of fully capturing the variety of movements found in the general population. Consequently, the concept of patient-specific fine-tuning appears beneficial for enhancing network performance on individual cases.

In this context, we investigate the impact of fine-tuning within a realistic scenario. Specifically, we assume that for a previously unseen subject, only seven movement examples belonging to a single class are available. This assumption aligns with a realistic clinical scenario, as it is common at the beginning of treatment to have only a few recorded executions of a movement, typically representing a consistent performance level.

To analyze the impact of augmented data in this fine-tuning scenario, we compared two experiments. In both, the initial model was pretrained on the original dataset with the target subject excluded, then fine-tuned on a small subject-specific set of five examples per class for training and two for validation. From the single class available for the target subject, these are its own real repetitions. The examples for the remaining classes were supplied from other sources, and it is here that the two experiments differ, along with the data used for pretraining:

\begin{itemize}
    \item TRTR-LOSO-FT (without augmented data): the model is pretrained on real data only, and the classes absent for the target subject are supplemented with real examples from the original dataset.
    \item TRATR-LOSO-FT (with augmented data): the model is pretrained on real and augmented data, and the absent classes are supplemented with augmented examples generated from the target subject's own repetitions.
\end{itemize}

In both experiments, fine-tuning was limited to the dense layers of the model, while the convolutional feature extraction layers remained fixed. Training was conducted for 100 epochs with an initial learning rate of $10^{-5}$. The same PyTorch ReduceLROnPlateau scheduler used for the main training reduced the learning rate by a factor of 0.5 when no improvement in validation loss was observed for 10 consecutive epochs (minimum learning rate $10^{-6}$). An early stopping patience of 100 epochs was set, effectively allowing training to complete for the full duration. The model checkpoint achieving the lowest validation loss (minimum delta $10^{-3}$) was retained for evaluation. Fine-tuning was applied to all subjects, including those who had already achieved a macro-F1 score of 1.0 under the LOSO evaluation, since subject-specific performance cannot be known a priori in a practical deployment scenario.

%%%%%%%%%%%%%%%%%%%%%%%%%%%%%%%%%%%%%%%%%%
\section{Results}
This section summarizes the experimental results. We first characterize the distribution of real and augmented samples in a t-SNE plot. We then assess the impact of augmented data on classification performance across different evaluation scenarios. Subsequently, we present the results of the two control experiments: the augmentation pipeline validation and the robustness analysis. This is followed by an analysis of subject-specific generalization under LOSO evaluation. Finally, we evaluate the benefits of incorporating augmented data into patient-specific fine-tuning strategies.

\subsection{Characteristics of real and augmented samples}
\begin{figure}[htbp]
\centering
\includegraphics[width=\linewidth]{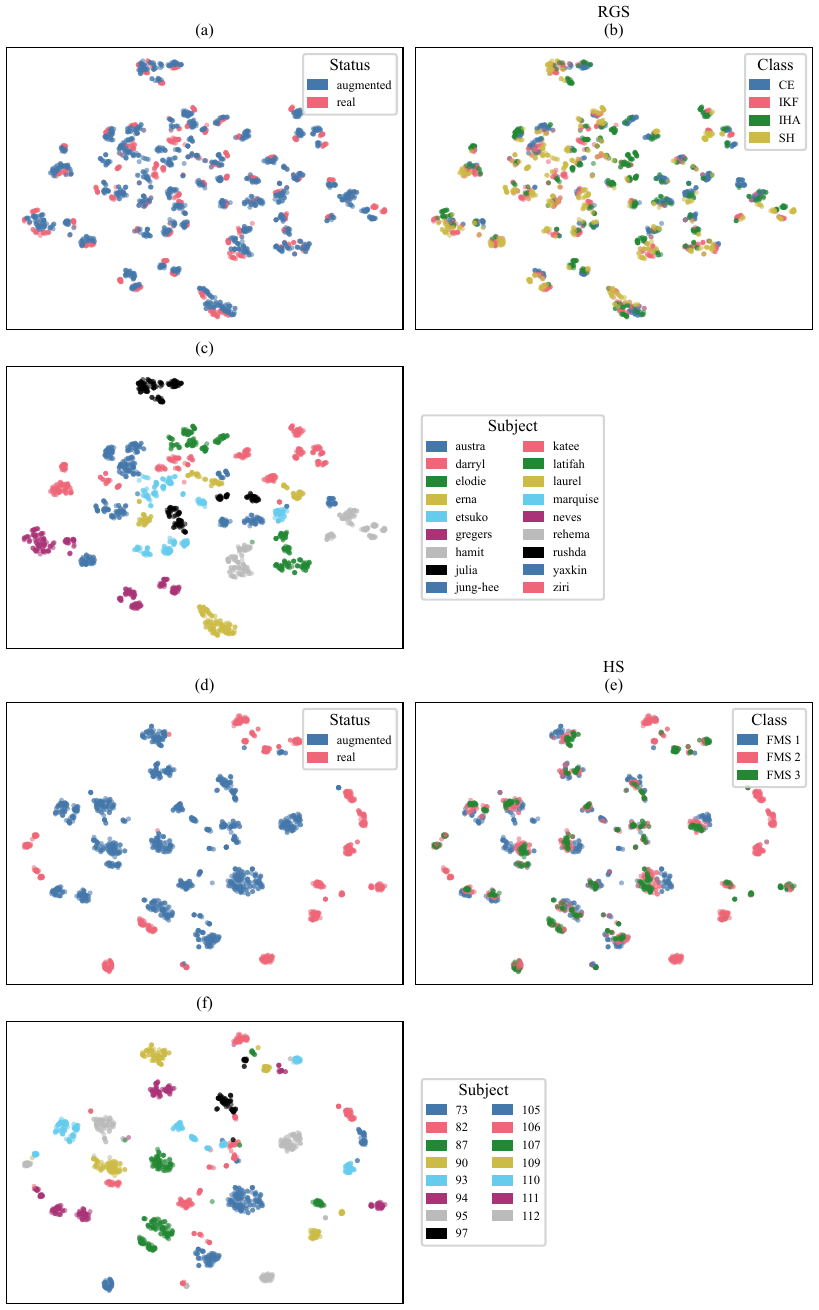}
\caption{T-SNE-based dimensionality reduction of augmented and real data for the RGS (a--c) and HS (d--f) datasets.\\
Within each dataset, the three panels show the same t-SNE embedding colored by a different criterion: (a,d) real versus augmented origin, (b,e) class label, and (c,f) subject.}
\label{fig:res_tsne}
\end{figure}

Figure~\ref{fig:res_tsne} presents, by way of example, the results of dimensionality reduction using t-SNE for both RGS and HS samples. The DS and RD datasets exhibit distributional behavior analogous to HS and RGS, respectively, and are therefore not separately visualized.

In the RGS dataset, augmented examples closely cluster around the corresponding real examples, suggesting that the augmentation process replicates the real-world movement characteristics. The embedded space reveals no clear separation between different classes. Notably, a prominent subject-specific clustering pattern is observed, with augmented examples consistently positioned within these subject clusters.

The HS dataset exhibits a more diverse distribution pattern. Here, multiple clusters contain predominantly or exclusively real or augmented data points. However, no distinct separation between the classes is observable here either. A clear distinction between subjects is still maintained.

\subsection{Classification performance}
Table~\ref{tab:2} provides a summary of the overall and class-specific classification performance, reported as mean $\pm$ standard deviation across the five cross-validation folds, for five distinct training and testing scenarios and all four datasets.

In the TRTR scenario, the RD (1.00) and RGS (0.99) datasets yielded near-perfect mean macro F1-scores. The DS dataset also achieved a high score of 0.93, while the HS dataset reached a noticeably lower value of 0.67. A closer inspection of the class-wise F1-scores reveals that the models trained on RD and RGS perform consistently well across all classes. For DS, class 3 was classified most accurately, followed by class 1, while performance for class 2 was comparatively lower. In contrast, the model trained on HS shows good classification results for classes 2 and 3, but a substantially lower performance for class 1, highlighting this class as a primary source of misclassification within this dataset. Standard deviations are low for RD, RGS, and DS. For HS, the standard deviation is markedly higher ($\pm$0.25 at the macro level, $\pm$0.40 for class 1).

In the TATR scenario, the RD dataset showed a notable performance drop, with the mean macro F1-score decreasing from 1.00 to 0.78. Class-wise, the TL and PRO categories were most affected, dropping to 0.68 and 0.69 respectively, while the SUP class remained comparatively robust at 0.91. Similarly, the RGS dataset declined from 0.99 to 0.83, with the IKF class showing the largest degradation (0.73), while IHA remained strong at 0.94. The DS dataset also declined noticeably to 0.81, with class 2 dropping most substantially to 0.61. The most pronounced degradation was observed on the HS dataset, where the mean macro F1-score fell to 0.59. Here, class 1 performed particularly poorly at 0.24, while classes 2 and 3 remained more robust. Standard deviations are low across all four datasets in this scenario.

In the TRTA scenario, a clear drop in performance was observed across all datasets. The RGS dataset was the most resilient, achieving a mean macro F1-score of 0.66, with class-wise scores ranging from 0.57 to 0.78. The DS dataset declined more substantially to 0.50, with all three classes performing similarly poorly. The RD dataset reached only 0.47, with the TL class showing the weakest result at 0.35. The HS dataset suffered the most substantial degradation, reaching a macro F1-score of 0.25. While class 3 still achieved a score of 0.51, the model completely failed to detect class 1, yielding an F1-score of 0.00. Standard deviations are low across all datasets.

For RD and RGS, TRATR remained within 0.01 of the TRTR baseline (0.99 and 0.98 vs.\ 1.00 and 0.99). For DS it exceeded the baseline (0.95 vs.\ 0.93), and for HS it produced a clear increase (0.77 vs.\ 0.67). Across all datasets, TRATR substantially outperformed both the augmented-only (TATR) and the real-on-augmented (TRTA) configurations. Standard deviations are low for RD, RGS, and DS. For HS, fold-to-fold variability remains elevated at the macro level ($\pm$0.10) and particularly for class 1 ($\pm$0.30), though both represent a reduction compared to the TRTR baseline.

\begin{table}[H]
\centering
\caption{Summary of the overall and class-specific performance across five training--testing configurations for the RD, RGS, DS, and HS datasets.\\
Results are reported as mean $\pm$ standard deviation across the five cross-validation folds. For each dataset, the subscripted class indices correspond to the following execution variants: RD: $1$=CE, $2$=TL, $3$=SUP, $4$=PRO; RGS: $1$=CE, $2$=IKF, $3$=IHA, $4$=SH; DS and HS: $1$--$3$ correspond to FMS scores 1, 2, and 3 respectively.}
\label{tab:2}
\setlength{\tabcolsep}{4.5pt}
\begin{tabular}{ccccccc}
\toprule
Exercise & Setup & Macro F1 & F1$_1$ & F1$_2$ & F1$_3$ & F1$_4$ \\
\midrule
\multirow[c]{5}{*}{\shortstack{RD\\{\small(CE / TL /}\\{\small SUP / PRO)}}} & TRTR & 1.00$\,\pm\,$0.00 & 0.99$\,\pm\,$0.01 & 1.00$\,\pm\,$0.01 & 1.00$\,\pm\,$0.01 & 1.00$\,\pm\,$0.00 \\
 & TRATR & 0.99$\,\pm\,$0.01 & 0.99$\,\pm\,$0.01 & 0.98$\,\pm\,$0.01 & 0.99$\,\pm\,$0.01 & 0.99$\,\pm\,$0.01 \\
 & TATR & 0.78$\,\pm\,$0.06 & 0.84$\,\pm\,$0.06 & 0.68$\,\pm\,$0.10 & 0.91$\,\pm\,$0.04 & 0.69$\,\pm\,$0.15 \\
 & TRTA & 0.47$\,\pm\,$0.02 & 0.53$\,\pm\,$0.02 & 0.35$\,\pm\,$0.02 & 0.59$\,\pm\,$0.03 & 0.42$\,\pm\,$0.09 \\
 & TRATA & 0.90$\,\pm\,$0.01 & 0.90$\,\pm\,$0.01 & 0.87$\,\pm\,$0.02 & 0.93$\,\pm\,$0.02 & 0.92$\,\pm\,$0.01 \\
\cline{1-7}
\multirow[c]{5}{*}{\shortstack{RGS\\{\small(CE / IKF /}\\{\small IHA / SH)}}} & TRTR & 0.99$\,\pm\,$0.01 & 0.99$\,\pm\,$0.01 & 0.98$\,\pm\,$0.01 & 1.00$\,\pm\,$0.01 & 0.99$\,\pm\,$0.01 \\
 & TRATR & 0.98$\,\pm\,$0.01 & 0.99$\,\pm\,$0.01 & 0.97$\,\pm\,$0.03 & 1.00$\,\pm\,$0.00 & 0.98$\,\pm\,$0.02 \\
 & TATR & 0.83$\,\pm\,$0.03 & 0.85$\,\pm\,$0.07 & 0.73$\,\pm\,$0.04 & 0.94$\,\pm\,$0.01 & 0.81$\,\pm\,$0.06 \\
 & TRTA & 0.66$\,\pm\,$0.03 & 0.67$\,\pm\,$0.04 & 0.57$\,\pm\,$0.03 & 0.61$\,\pm\,$0.05 & 0.78$\,\pm\,$0.05 \\
 & TRATA & 0.92$\,\pm\,$0.01 & 0.90$\,\pm\,$0.02 & 0.85$\,\pm\,$0.03 & 0.96$\,\pm\,$0.01 & 0.95$\,\pm\,$0.02 \\
\cline{1-7}
\multirow[c]{5}{*}{\shortstack{DS\\{\small(FMS}\\{\small 1 / 2 / 3)}}} & TRTR & 0.93$\,\pm\,$0.03 & 0.95$\,\pm\,$0.02 & 0.85$\,\pm\,$0.06 & 1.00$\,\pm\,$0.00 &  \\
 & TRATR & 0.95$\,\pm\,$0.01 & 0.97$\,\pm\,$0.00 & 0.89$\,\pm\,$0.02 & 1.00$\,\pm\,$0.00 &  \\
 & TATR & 0.81$\,\pm\,$0.01 & 0.84$\,\pm\,$0.01 & 0.61$\,\pm\,$0.02 & 0.98$\,\pm\,$0.03 &  \\
 & TRTA & 0.50$\,\pm\,$0.04 & 0.58$\,\pm\,$0.03 & 0.58$\,\pm\,$0.06 & 0.34$\,\pm\,$0.11 &  \\
 & TRATA & 0.98$\,\pm\,$0.01 & 0.97$\,\pm\,$0.01 & 0.97$\,\pm\,$0.01 & 0.99$\,\pm\,$0.00 &  \\
\cline{1-7}
\multirow[c]{5}{*}{\shortstack{HS\\{\small(FMS}\\{\small 1 / 2 / 3)}}} & TRTR & 0.67$\,\pm\,$0.25 & 0.51$\,\pm\,$0.40 & 0.71$\,\pm\,$0.36 & 0.78$\,\pm\,$0.11 &  \\
 & TRATR & 0.77$\,\pm\,$0.10 & 0.55$\,\pm\,$0.30 & 0.90$\,\pm\,$0.01 & 0.85$\,\pm\,$0.02 &  \\
 & TATR & 0.59$\,\pm\,$0.07 & 0.24$\,\pm\,$0.18 & 0.81$\,\pm\,$0.02 & 0.72$\,\pm\,$0.05 &  \\
 & TRTA & 0.25$\,\pm\,$0.04 & 0.00$\,\pm\,$0.00 & 0.26$\,\pm\,$0.13 & 0.51$\,\pm\,$0.02 &  \\
 & TRATA & 0.95$\,\pm\,$0.01 & 0.99$\,\pm\,$0.00 & 0.93$\,\pm\,$0.02 & 0.95$\,\pm\,$0.01 &  \\
\cline{1-7}
\bottomrule
\end{tabular}
\end{table}

\subsection{Augmentation pipeline validation}
\label{sec:pipeline_val}

Table~\ref{tab:pipeline_val} reports macro F1-scores for the pipeline validation experiment, in which processed but unmodified repetitions replace augmented data in the TPTR and TRTP paradigms.

\begin{table}[H]
\centering
\caption{Macro F1-scores for the augmentation pipeline validation experiment.\\
TPTR and TRTP denote conditions in which processed but unmodified repetitions were used in place of augmented data. Results are reported as mean $\pm$ standard deviation across the five cross-validation folds.}
\label{tab:pipeline_val}
\begin{tabular}{lccc}
\toprule
Exercise & TRTR & TPTR & TRTP \\
\midrule
RD  & 1.00 $\pm$ 0.00 & 1.00 $\pm$ 0.00 & 1.00 $\pm$ 0.00 \\
RGS & 0.99 $\pm$ 0.01 & 0.99 $\pm$ 0.01 & 0.99 $\pm$ 0.01 \\
DS  & 0.93 $\pm$ 0.03 & 0.94 $\pm$ 0.02 & 0.94 $\pm$ 0.01 \\
HS  & 0.67 $\pm$ 0.25 & 0.75 $\pm$ 0.14 & 0.71 $\pm$ 0.19 \\
\bottomrule
\end{tabular}
\end{table}

For RD and RGS, scores in both pipeline conditions match the TRTR baseline exactly. DS shows values of 0.94~$\pm$~0.02 and 0.94~$\pm$~0.01, within one standard deviation of the TRTR baseline of 0.93~$\pm$~0.03. For HS, the pipeline conditions yield 0.75~$\pm$~0.14 and 0.71~$\pm$~0.19. Compared to 0.67~$\pm$~0.25 under TRTR, the standard deviations remain elevated, consistent with the high fold-to-fold variability observed for HS throughout.

\subsection{Robustness analysis}
\label{sec:robustness}

Table~\ref{tab:robustness} reports macro F1-scores on augmented test data for the TRTA and TRATA conditions. TRTA trains on real data only, whereas TRATA uses the same combined real and augmented training set as TRATR. Under TRTA, macro F1-scores range from 0.25 (HS) to 0.66 (RGS). Under TRATA, they range from 0.90 (RD) to 0.98 (DS). Standard deviations are low across all datasets in both conditions.

\begin{table}[H]
\centering
\caption{Macro F1-scores on augmented test data for a classifier trained on real data only (TRTA) and a classifier trained on the combined real and augmented set (TRATA).\\
Results are reported as mean $\pm$ standard deviation across the five cross-validation folds.}
\label{tab:robustness}
\begin{tabular}{lcc}
\toprule
Exercise & TRTA & TRATA \\
\midrule
RD  & 0.47 $\pm$ 0.02 & 0.90 $\pm$ 0.01 \\
RGS & 0.66 $\pm$ 0.03 & 0.92 $\pm$ 0.01 \\
DS  & 0.50 $\pm$ 0.04 & 0.98 $\pm$ 0.01 \\
HS  & 0.25 $\pm$ 0.04 & 0.95 $\pm$ 0.01 \\
\bottomrule
\end{tabular}
\end{table}

\subsection{Generalization to unseen subjects}
\begin{figure}[H]
\centering
\includegraphics[width=\linewidth]{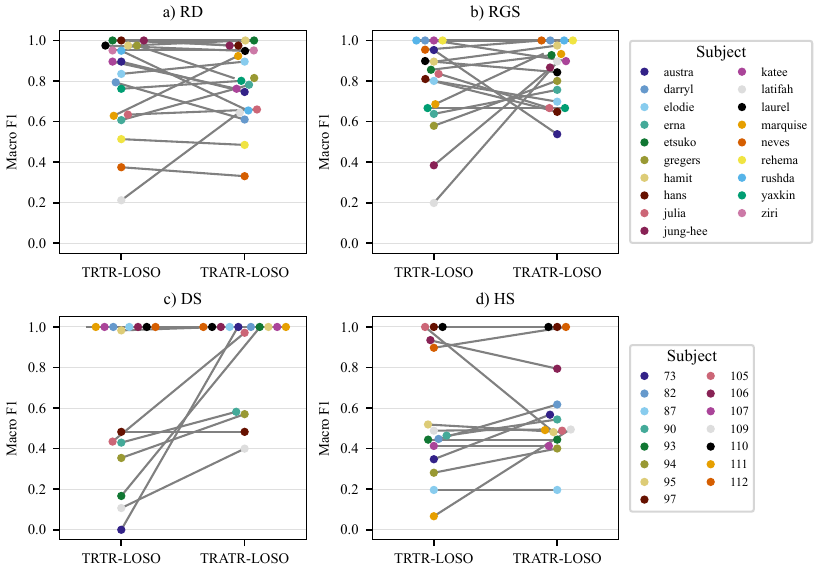}
\caption{Comparison of subject-wise performance (LOSO) on real data between a classifier trained on exclusively real data (TRTR-LOSO) and a classifier trained on a combination of real and augmented data (TRATR-LOSO).\\
This evaluation is performed for four datasets: RD (a), RGS (b), DS (c), and HS (d).}
\label{fig:res_loso}
\end{figure}

Table~\ref{tab:loso_overview} summarizes the mean macro F1-scores achieved under 
the TRTR-LOSO and TRATR-LOSO conditions for each exercise. Standard deviations 
are high across all datasets and conditions, indicating substantial variability 
in per-subject performance. Figure~\ref{fig:res_loso} provides a subject-level 
breakdown of these differences.

\begin{table}[H]
\centering
\caption{Mean macro F1-scores across all subjects for the TRTR-LOSO and TRATR-LOSO
conditions, reported as mean $\pm$ standard deviation across subjects.}
\label{tab:loso_overview}
\begin{tabular}{lcc}
\toprule
Exercise & TRTR-LOSO & TRATR-LOSO \\
\midrule
RD  & 0.79 $\pm$ 0.23 & 0.79 $\pm$ 0.19 \\
RGS & 0.79 $\pm$ 0.23 & 0.84 $\pm$ 0.15 \\
DS  & 0.66 $\pm$ 0.39 & 0.87 $\pm$ 0.23 \\
HS  & 0.57 $\pm$ 0.32 & 0.60 $\pm$ 0.24 \\
\bottomrule
\end{tabular}
\end{table}

For the RD dataset, the mean macro F1-score remained nearly unchanged when augmented data were included (0.79 under both conditions). At the subject level, 7 of 19 subjects improved, 10 declined, and 2 remained unchanged (Figure~\ref{fig:res_loso}~a). Several subjects with low baseline performance showed substantial gains (e.g., latifah from 0.21 to 0.64, marquise from 0.63 to 0.92), while some previously well-performing subjects declined (e.g., rushda from 0.95 to 0.65, darryl from 0.79 to 0.61).

For the RGS dataset, the inclusion of augmented data increased the mean macro F1-score from 0.79 to 0.84. At the subject level, 8 of 18 subjects improved, 6 declined, and 4 remained unchanged (Figure~\ref{fig:res_loso}~b). The strongest gains occurred for subjects with low baselines (laurel from 0.20 to 0.90, katee from 0.38 to 0.87). One subject showed a pronounced decline (austra from 0.95 to 0.54).

In the DS dataset, the mean macro F1-score increased from 0.66 to 0.87. At the subject level, 6 of 15 subjects improved and 9 remained unchanged, with no subject declining (Figure~\ref{fig:res_loso}~c). Subjects 73 (0.00 to 1.00), 93 (0.17 to 1.00), and 105 (0.43 to 0.97) showed the largest gains. All subjects who already achieved high scores under TRTR-LOSO maintained their performance.

Results for the HS dataset were more heterogeneous. The mean macro F1-score increased slightly from 0.57 to 0.60. At the subject level, 6 of 15 subjects improved, 3 declined, and 6 remained unchanged (Figure~\ref{fig:res_loso}~d). Subject 111 showed the largest gain (0.07 to 0.49), while subject 105 exhibited a pronounced decline (1.00 to 0.49). The majority of subjects remained in the 0.20 to 0.62 range under both conditions. 

\subsection{Patient-specific fine-tuning}
\begin{figure}[H]
\centering
\includegraphics[width=\linewidth]{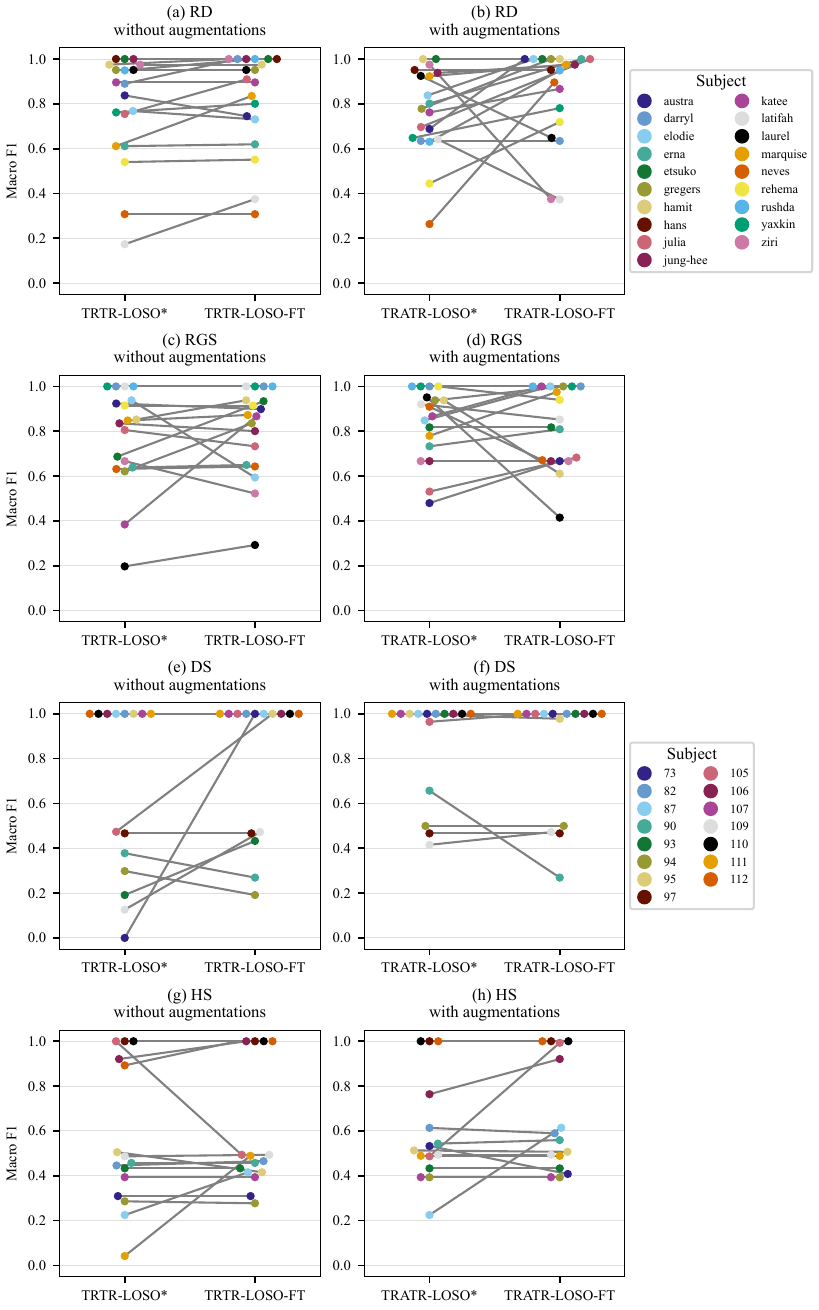}
\caption{Subject-wise macro F1-scores before and after patient-specific fine-tuning for the TRTR-LOSO-FT (a, c, e, g) and TRATR-LOSO-FT (b, d, f, h) conditions, shown for RD, RGS, DS, and HS respectively.\\
Each panel compares the pre-fine-tuning baseline (TRTR-LOSO* or TRATR-LOSO*) against the post-fine-tuning score for each subject.}
\label{fig:res_psft}
\end{figure}

To evaluate the subject-specific impact of fine-tuning, we compared classification performance before and after applying a fine-tuning step. This analysis was performed separately for models trained without and with augmented samples. For each of the four datasets, models were first trained in a LOSOCV setting and subsequently fine-tuned using a subset of real data or a combination of real and augmented data from the left-out subject. Since the real examples used for fine-tuning were excluded from the test set, the pre-fine-tuning baselines reported here differ slightly from the values in Table~\ref{tab:loso_overview} and are therefore denoted TRTR-LOSO* and TRATR-LOSO*. Table~\ref{tab:ft_overview} provides an exercise-level overview of the mean macro F1-scores before and after fine-tuning under both conditions. As in the LOSO evaluation, standard deviations are high across all exercises and conditions, reflecting substantial between-subject variability. Figure~\ref{fig:res_psft} shows the corresponding subject-level results.

\begin{table}[htbp]
\centering
\caption{Mean macro F1-scores before and after patient-specific fine-tuning for
each exercise, reported as mean $\pm$ standard deviation across subjects.\\
TRTR-LOSO* and TRATR-LOSO* denote the pre-fine-tuning baselines with the
fine-tuning examples excluded from the test set. TRTR-LOSO-FT and TRATR-LOSO-FT denote
the corresponding post-fine-tuning scores.}
\label{tab:ft_overview}
\begin{tabular}{lcccc}
\toprule
Exercise & TRTR-LOSO* & TRTR-LOSO-FT & TRATR-LOSO* & TRATR-LOSO-FT \\
\midrule
RD  & 0.79 $\pm$ 0.24 & 0.82 $\pm$ 0.22 & 0.77 $\pm$ 0.20 & 0.85 $\pm$ 0.21 \\
RGS & 0.78 $\pm$ 0.22 & 0.81 $\pm$ 0.20 & 0.84 $\pm$ 0.16 & 0.82 $\pm$ 0.18 \\
DS  & 0.66 $\pm$ 0.39 & 0.79 $\pm$ 0.32 & 0.87 $\pm$ 0.23 & 0.85 $\pm$ 0.27 \\
HS  & 0.56 $\pm$ 0.32 & 0.58 $\pm$ 0.27 & 0.59 $\pm$ 0.24 & 0.65 $\pm$ 0.25 \\
\bottomrule
\end{tabular}
\end{table}

For the RD dataset, fine-tuning with augmented data achieved a higher mean macro F1-score (0.85) than fine-tuning without augmented data (0.82). Without augmented data, fine-tuning produced a moderate improvement over the baseline (0.79 to 0.82), with 7 of 19 subjects improving, 2 declining, and 10 remaining unchanged (Figure~\ref{fig:res_psft}~a). Julia (0.75 to 0.91) and marquise (0.61 to 0.84) showed the largest gains. With augmented data, fine-tuning yielded a larger improvement (0.77 to 0.85), with 12 of 19 subjects improving and 6 reaching a macro F1-score of 1.00 (Figure~\ref{fig:res_psft}~b). However, 3 subjects showed pronounced drops, in particular ziri (0.97 to 0.38) and latifah (0.64 to 0.37).

For the RGS dataset, fine-tuning with and without augmented data yielded similar final scores (0.82 and 0.81). Without augmented data, fine-tuning improved 6 of 18 subjects, while 5 declined and 7 remained unchanged (Figure~\ref{fig:res_psft}~c). Katee showed the largest gain (0.38 to 0.87). With augmented data, the pre-fine-tuning baseline was already higher (0.84), and fine-tuning did not improve the mean further (0.82). At the subject level, 7 improved and 5 declined (Figure~\ref{fig:res_psft}~d), with laurel (0.95 to 0.41) and hamit (0.94 to 0.61) showing the largest drops.

In the DS dataset, fine-tuning with augmented data produced a higher final score (0.85) than without (0.79). Without augmented data, fine-tuning improved the mean from 0.66 to 0.79, with 4 of 15 subjects improving and 2 declining (Figure~\ref{fig:res_psft}~e). Subjects 73 (0.00 to 1.00) and 105 (0.47 to 1.00) reached perfect scores. With augmented data, the pre-fine-tuning baseline was already at 0.87, and fine-tuning produced only minor changes (0.85), with 2 subjects improving and 2 declining (Figure~\ref{fig:res_psft}~f). Subject 90 declined in both conditions.

The HS dataset showed the clearest advantage for augmentation-based fine-tuning (0.65 vs. 0.58). Without augmented data, fine-tuning had a marginal effect on the mean (0.56 to 0.58), with 4 subjects improving and 2 declining (Figure~\ref{fig:res_psft}~g). Subject 111 showed the largest gain (0.04 to 0.49), while subject 105 declined from 1.00 to 0.49. With augmented data, fine-tuning improved the mean from 0.59 to 0.65, with 3 subjects improving (Figure~\ref{fig:res_psft}~h). Subject 105 (0.49 to 0.99) and subject 87 (0.22 to 0.61) showed the largest gains.

%%%%%%%%%%%%%%%%%%%%%%%%%%%%%%%%%%%%%%%%%%

\section{Discussion}
In this work, we introduced a novel approach for data augmentation in IMU-based exercise evaluation, specifically tailored for physiotherapeutic exercises. In the following we want to discuss our main contributions and future implications.

% Automatic Labeling
\subsection{Automatic labeling quality}
Our automatic labeling pipeline assigns class labels by evaluating predefined decision
rules on inverse-kinematics-derived metrics. Each evaluation criterion uses fixed
numeric thresholds to guarantee consistent and reproducible labeling. The pipeline
achieves $GM_{F_1}$ scores of 0.88 (RD), 0.83 (RGS), 0.84 (DS), and 0.75 (HS).
These values indicate that the rule-based classification closely approximates the
labels present in the real dataset, but cannot fully reproduce them. 

We attribute this discrepancy to the interaction between two factors. First,
movement quality is more accurately described as a continuum than as a set of
discrete classes, so that a substantial proportion of repetitions falls near the
boundary between adjacent classes. For DS and HS, this is directly evidenced by the
ambiguous repetitions shown in Figure~\ref{fig:ambivalent_percentage}. For RD and
RGS, no equivalent annotation is available, but the imperfect $GM_{F_1}$ scores
indicate that boundary-proximal cases are present in these datasets as well.
Second, the threshold optimization searches for fixed numeric cut-offs that
separate the data as well as possible. Within the dense cluster of boundary-proximal
repetitions, multiple threshold configurations yield comparable $GM_{F_1}$ scores
while assigning some of these repetitions to different classes. The optimizer
therefore selects a decision boundary that lies close to, but not exactly at, the
one implicit in the original labels.

As a consequence, augmented and real data can carry slightly different decision
boundaries when used as independent training sets. Augmented examples are labeled
according to the optimized thresholds, while real examples carry their original
expert labels. A network trained on either set learns the corresponding boundary,
and the implications of this for classification performance are discussed in the
following sections. Future work should extend the automatic labeling framework to
represent label uncertainty explicitly, for example by providing a label
distribution instead of a single hard label.

However, it is crucial to emphasize that the automated labeling system should not be viewed as a replacement for neural network-based classification. A key limitation is the extensive data requirements of the automated labeling, which relies on inverse kinematics calculations to determine joint angles, positions, and related metrics. These calculations necessitate at least one IMU per moving segment, making it impossible to reduce the number of sensors for complex exercises. Currently, our neural network based classification also employs a high number of IMUs. However, ongoing research within our group specifically investigates methods to reduce the required number of IMUs for neural network-based classification, aiming to enhance practical applicability in clinical settings.

\subsection{Scope and limitations}
Our data augmentation approach successfully generates examples that respect physiological joint-range limits, and can represent both identical and distinct movement classes. This capability is particularly beneficial for addressing imbalanced datasets, as it enables the targeted generation of rare or specific movement patterns that are typically underrepresented. Importantly, the method is not inherently restricted to the presented exercises. Rather, it can be extended to additional exercises, provided that at least a minimal number of real repetitions are recorded using a suitable IMU setup and sufficient domain knowledge regarding correct execution and variations is available. In physiotherapeutic contexts, such requirements are commonly met, making this approach particularly suitable for clinical applications.

However, extending this method to new exercises does require adjustments. Specifically, the automatic labeling must be adapted to reflect new evaluation criteria, necessitating the definition of new decision thresholds and distributions for each Euler angle, derived either from real data, domain knowledge, or a combination thereof.

The computational demands of our augmentation process remain comparatively low. Unlike approaches based on generative neural networks like GANs, our approach does not require substantial GPU resources. Indeed, the computational time needed for generating an augmented example approximately equals the duration of the original exercise, a highly acceptable cost for practical applications.

One limitation of the procedure presented is the lack of subject-specific scaling of the biomechanical model. This was omitted because the required anthropometric data for the DS and HS datasets are not available. Introducing individual scaling could potentially result in more personalized augmented examples, although it remains unclear whether this increased subject specificity would positively or negatively impact the overall generalization of neural network models. Furthermore, the biomechanical model currently used offers limited degrees of freedom in the torso region and therefore represents a simplified approximation of human anatomy. Thus, for exercises predominantly involving torso movements, a more detailed biomechanical representation of this region would be advisable to enhance the realism of generated examples. 

Additionally, while our pipeline enforces kinematic limits, a firmer guarantee of biomechanical plausibility could potentially increase the realism of the generated examples and would require incorporating kinetics. In particular, explicitly accounting for external and contact forces and estimating net joint moments would allow enforcing torque and force bounds in the augmented repetitions. Implementing such kinetic constraints is feasible in OpenSim via inverse-dynamics or muscle-driven simulations, but it typically requires datasets that provide external and contact forces (e.g., force-plate or instrumented insole measurements). Recent optimal control approaches~\citep{nitschke_estimating_2024} estimate ground reaction forces from inertial data alone through an embedded contact model, removing the force-plate requirement but at the cost of hours of computation per trial. Either route would substantially increase the complexity of the augmentation pipeline. Therefore, while kinetic modeling could further enhance biomechanical realism, it would come at the cost of greater experimental or computational complexity and reduced applicability in real-world or clinical data collection scenarios.

\modified{A further limitation concerns the classifier used to evaluate the augmented data. Its network training hyperparameters were not optimized but fixed across all experiments, as described in the Methods. A dedicated hyperparameter optimization, performed separately per dataset, could raise the absolute classification scores, in particular for the more challenging HS dataset. Since our objective is to isolate the effect of augmentation rather than to maximize classifier performance, we leave such optimization to future work.}

\subsection{Characteristics of real and augmented samples}
The t-SNE analysis revealed structural similarities between real and augmented 
samples. For the RGS dataset, real and augmented examples formed shared clusters, 
indicating that the fundamental properties of the exercise repetitions were 
preserved. The HS dataset showed a more distinct separation between real and 
augmented examples, with fewer mixed clusters. This does not necessarily imply 
substantial discrepancies between the two sample types. Rather, it may indicate 
that the augmented HS samples encompass a broader range of movement variations 
than the real data, which themselves exhibit only limited variability in execution.

The frequent occurrence of clusters encompassing examples from multiple classes 
reflects the complexity of class separation within the dataset, and supports the 
use of classification methods such as the neural network applied in this study, 
which are capable of capturing nuanced differences between movement qualities.

The formation of subject-specific clusters, populated by both real and augmented 
examples, indicates that the augmentation method preserves patient-specific 
movement characteristics. This property is particularly relevant for fine-tuning 
scenarios, where augmented data can increase the amount and variability of the 
limited subject-specific training examples available.

\subsection{Classification performance}
Across all four exercises, the evaluation paradigms follow the pattern\\ \mbox{TRATR $\approx$ TRTR $>$ TATR $>$ TRTA}, with TRATR matching the TRTR baseline for RD, RGS, and DS and clearly exceeding it for HS.

% TRATR >= TRTR
That TRATR matches or exceeds TRTR indicates that augmented samples do not displace real data patterns but extend the existing variance in the training set. For RD and RGS, the differences between TRATR and TRTR are negligible and likely attributable to random variation, as the respective TRTR baselines already operate near saturation. A similar observation holds for DS, where the improvement remains small (0.93 to 0.95), as the baseline already achieves high classification performance. 

For HS, where the TRTR baseline leaves the most room for improvement, augmented training data produce the most pronounced gain (0.67 to 0.77). We attribute this to the augmented examples filling underrepresented regions of the HS feature space, exposing the model to execution variants that are sparse in the real recordings. The same mechanism is visible in the fold-to-fold variance. Under TRTR, HS carries the highest variance of all datasets ($\pm$0.25 at the macro level, $\pm$0.40 for class 1), a consequence of its small per-class sample size: when a class contributes only a few test instances to a fold, a single misclassification shifts its F1-score substantially, and the macro average inherits this sensitivity. Adding augmented data lowers the variance to $\pm$0.10 at the macro level and $\pm$0.30 for class 1, indicating that the broader coverage stabilizes training and reduces the model's sensitivity to the particular fold composition.

% Performance TATR & TRTA vs TRTR
Training exclusively on augmented data (TATR) consistently yields lower performance on real test data across all four exercises compared to TRTR (RD: 1.00 to 0.78, RGS: 0.99 to 0.83, DS: 0.93 to 0.81, HS: 0.67 to 0.59), indicating a systematic difference between the two data types, consistent with the low fold-to-fold variance in this condition. One possible cause is that the augmentation pipeline itself introduces artifactual signal characteristics absent from real recordings. 

The pipeline validation results rule out this explanation (TPTR and TRTP, see Section~\ref{sec:pipeline_val}): passing real repetitions through the complete pipeline without scaling or offset modifications yields performance on par with the TRTR baseline across all four datasets, confirming that the processing chain does not introduce a systematic distribution shift.

Instead, we attribute the performance drop in TATR to the slightly different decision boundary carried by the augmented labels, as discussed in the context of the automatic labeling pipeline. A network trained exclusively on augmented data learns this boundary, and examples that lie on the opposite side of the corresponding real-data boundary are classified differently than the real labels prescribe. By symmetry, the same mechanism applies to TRTA: a classifier trained on real data does not align with the augmented boundary, causing analogous misclassifications on augmented test data. TRATR incorporates both distributions during training and thereby stabilizes the decision boundary.

% Robustness
The TRATA condition shows the robustness benefit of adding augmented data to the training set (Table~\ref{tab:robustness}). A classifier trained only on real data classifies augmented test data poorly, whereas the combined training set yields high macro F1-scores on augmented test data as well. Combined with the TRATR results, this indicates that joint training performs well on both real and augmented test data without a meaningful trade-off between the two. As discussed above, diverging decision boundaries account for part of the TRTA performance drop, but we hypothesize that they are not the sole factor. Augmented examples cover a broader and more uniform region of the feature space than real recordings: real data form dense, subject-specific clusters around modal movement patterns, whereas augmented data extend coverage toward the margins of the class distributions. A model trained exclusively on real data is unlikely to have encountered movement variations near these margins and may therefore lack the decision boundaries required to classify them correctly.

This observation carries an implication beyond the experimental setting. Real recordings from unseen subjects or novel movement contexts may similarly populate underrepresented regions of the feature space. A model trained on the broader distribution provided by augmented data will have been exposed to such variations during training, which may translate into a genuine generalization advantage in deployment. Confirming this effect will require evaluation on additional datasets that systematically cover such out-of-distribution variations.

\subsection{Generalization to unseen subjects}

Incorporating augmented data improved the mean macro F1-score for three of the four exercises and left it unchanged for RD (Table~\ref{tab:loso_overview}). These means hide substantial variation between subjects. The same imbalance mechanism discussed above drives this variation: under LOSO, a held-out subject often contributes repetitions from only a subset of the classes, so that individual results swing widely. Most subjects gained moderately or substantially from augmentation, with Subject 73 in the DS dataset the clearest case. Because Subject 73 is the only contributor of class 3 examples (Figure~\ref{fig:rep_per_subject_and_label}~c), removing it under LOSO leaves no class 3 training data. Augmented examples filled this gap and raised its macro F1-score from 0.00 to 1.00. A number of subjects, however, declined when augmented data were included.

We attribute these declines to the absence of the affected subjects' movement patterns from the training set. The t-SNE analysis showed that augmented examples preserve the subject-specific characteristics of their source repetitions, so they add no new information about an unseen subject. This motivates patient-specific fine-tuning as a complementary strategy for subjects who do not benefit from augmentation alone.

\subsection{Patient-specific fine-tuning}

Comparing the TRATR-LOSO-FT and TRTR-LOSO-FT with the TRATR-LOSO* and TRTR-LOSO* scores in Table~\ref{tab:ft_overview}, fine-tuning improved the mean macro F1-score over the respective baseline for all exercises, with two exceptions: for RGS and DS with augmented data, the mean did not increase (RGS: 0.84 to 0.82, DS: 0.87 to 0.85). In both cases, the TRATR-LOSO* baseline was already substantially above the corresponding TRTR-LOSO* baseline (RGS: 0.84 vs. 0.78, DS: 0.87 vs. 0.66), indicating that the augmented training data had already captured much of the achievable performance gain prior to fine-tuning. Across all four exercises, TRATR-LOSO-FT produced equal or higher final macro F1-scores than TRTR-LOSO-FT, indicating that fine-tuning with augmented data consistently outperformed fine-tuning without.

At the subject level, TRATR-LOSO-FT results were mixed (Figure~\ref{fig:res_psft}). The majority of subjects benefited from or were unaffected by fine-tuning, but a number of individual subjects showed pronounced performance declines across all four exercises.

We hypothesize that these declines are driven by label ambiguity. The network is in principle capable of adapting its decision boundaries to subject- and class-specific movement characteristics, as demonstrated by the high performance in the non-LOSO 5-fold CV setting (Table~\ref{tab:2}), where all subjects contribute training examples. When real examples of a class are available for a subject during fine-tuning, the network can adapt to that subject's specific patterns, including ambiguous repetitions near the class boundary. When such examples are absent and replaced by augmented data, classification of ambiguous repetitions may deteriorate, since the augmented data preserve the source subject's characteristics rather than introducing the target subject's specific boundary patterns.

To probe this hypothesis post hoc, we conducted an exploratory analysis for the DS and HS datasets, where per-repetition annotator agreement data are available. For each class-subject combination, we determined (a) whether the fine-tuning training set contained real examples of that class for the respective subject, and (b) the proportion of ambiguously labeled repetitions among the test examples of that class for that subject. Table~\ref{tab:ambivalence_ft} summarizes the results, grouping combinations by both criteria.

\begin{table}[H]
\centering
\caption{Exploratory analysis of class-level $F_1$ changes after fine-tuning per class-subject combination in the DS and HS datasets (TRATR-LOSO-FT condition).\\
Combinations are grouped by whether real examples of the class were available in the fine-tuning set and by the ambiguity level of the test examples. $n$ is the number of combinations per group. Improved, Unchanged, and Declined count the combinations whose $F_1$ increased, stayed identical, or decreased. $\overline{|\Delta F_1|}$ is the mean magnitude of decline over the declined combinations only.}
\label{tab:ambivalence_ft}
\setlength{\tabcolsep}{4pt}
\begin{tabular}{llccccc}
\toprule
\multirow{2}{*}{\shortstack[l]{Real examples\\in training set}} & \multirow{2}{*}{\shortstack[l]{Ambiguity\\level}} & \multirow{2}{*}{$n$} & \multicolumn{3}{c}{Class-level $F_1$ change} & \multirow{2}{*}{\shortstack{Mean decline\\$\overline{|\Delta F_1|}$}} \\
\cmidrule(lr){4-6}
 & & & Improved & Unchanged & Declined & \\
\midrule
\multirow{3}{*}{Available} & Low (0--25\%)     & 21 & 5 (24\%)  & 15 (71\%) & 1 (5\%)   & 0.02 \\
                           & Medium (26--60\%) & 8  & 3 (38\%)  & 5 (62\%)  & 0 (0\%)   & --   \\
                           & High ({>}60\%)    & 1  & 0 (0\%)   & 1 (100\%) & 0 (0\%)   & --   \\
\midrule
\multirow{3}{*}{Absent}    & Low (0--25\%)     & 4  & 1 (25\%)  & 3 (75\%)  & 0 (0\%)   & --   \\
                           & Medium (26--60\%) & 4  & 0 (0\%)   & 2 (50\%)  & 2 (50\%)  & 0.12 \\
                           & High ({>}60\%)    & 9  & 0 (0\%)   & 5 (56\%)  & 4 (44\%)  & 0.26 \\
\bottomrule
\end{tabular}
\end{table}

Among class-subject combinations where real training examples were available, only 1 of 30 across all ambiguity levels showed a performance decline, with a magnitude of only 0.02 that can be attributed to random variation. Among combinations without real training examples, the decline rate increased with the ambiguity level: 0 of 4 at low ambiguity, 2 of 4 at medium ambiguity (mean decline 0.12), and 4 of 9 at high ambiguity (mean decline 0.26). The absence of declines at low ambiguity without real examples indicates that the lack of real training data alone does not cause performance drops. Rather, the combination of absent real examples and high label ambiguity appears to drive the observed declines.

For RD and RGS, no per-repetition ambiguity annotation is available, so this analysis cannot be repeated directly. However, these exercises belong to the same domain of movement quality assessment as DS and HS. It is inherent to this domain that individual repetitions can fall near the decision boundaries between classes and are therefore effectively ambiguous. The occurrence of such boundary-proximal examples in the RD and RGS datasets is plausible, but cannot be verified due to the absence of corresponding annotations.

In principle, the augmentation pipeline can generate examples near class boundaries, but reliably reproducing the specific ambiguity patterns of an individual subject would require access to that subject's test data, which constitutes data leakage. 

To further investigate this effect and improve performance in the presence of label ambiguity, two directions appear promising: adopting training methods that explicitly account for label uncertainty, such as label distribution learning approaches~\citep{gaoDeepLabelDistribution2017a}, and developing methods to reliably identify and flag ambiguous examples in both real and augmented data.

\subsection{Clinical applicability}
The proposed augmentation approach has potential use for clinical workflows in physiotherapy and rehabilitation. By generating labeled IMU trajectories across different performance levels, it can expand scarce clinical datasets for training robust assessment models and it can personalize pretrained models to individual patients. A practical workflow is as follows. At treatment onset, a therapist records and rates a small number of patient repetitions. Our pipeline then augments these movements and labels the additional variants automatically. The classifier is fine-tuned on this patient-specific set and subsequently deployed for home use to provide real-time execution feedback. The same predictions can be reviewed retrospectively as a monitoring tool to document progress and to support therapist decision-making, for example when adjusting exercise difficulty or focus. In this way, augmentation combined with few-shot fine-tuning addresses the common problem of poor generalization to unseen patients while keeping data requirements and computation modest enough for routine use.

%%%%%%%%%%%%%%%%%%%%%%%%%%%%%%%%%%%%%%%%%%
\section{Conclusion}

We presented a musculoskeletal simulation-based data augmentation method for IMU-based exercise evaluation that operates on IMU recordings alone, enforces anatomically plausible joint-range limits, and labels augmented repetitions automatically by combining inverse kinematics parameters with a knowledge-based evaluation strategy. Across four datasets of increasing complexity, adding augmented data to real training data matched or improved classification performance, with the largest gains on the imbalanced HS dataset and under leave-one-subject-out evaluation, including subjects whose training data lacked an entire movement class. Augmented data further enabled patient-specific fine-tuning from few real repetitions. The magnitude of these gains depended on dataset properties, in particular class balance and label ambiguity: subjects benefited most when real examples of a class were unavailable and label ambiguity was low, while the combination of absent real data and high ambiguity limited the benefit. Remaining limitations include the absence of subject-specific model scaling, a simplified torso representation, the reliance on kinematic rather than kinetic constraints, and an automatic labeling step that assigns single hard labels. Future work should represent label ambiguity explicitly and evaluate the generalization benefit on datasets that systematically cover out-of-distribution movement variations.

%%%%%%%%%%%%%%%%%%%%%%%%%%%%%%%%%%%%%%%%%%
\vspace{6pt} 

%%%%%%%%%%%%%%%%%%%%%%%%%%%%%%%%%%%%%%%%%%
%% optional
%\supplementary{The following supporting information can be downloaded at \linksupplementary{s1}, Figure S1: title; Table S1: title; Video S1: title.}

% Only used for preprtints:
% \supplementary{The following supporting information can be downloaded at the website of this paper posted on \href{https://www.preprints.org/}{Preprints.org}.}

%%%%%%%%%%%%%%%%%%%%%%%%%%%%%%%%%%%%%%%%%%
\authorcontributions{Conceptualization, A.S., H.O. and M.M.; methodology, A.S., H.O. and M.M.; software, A.S. and M.M.; validation, A.S. and M.M.; formal analysis, A.S. and M.M.; investigation, A.S. and M.M.; resources, M.M.; data curation, A.S.; writing---original draft preparation, A.S.; writing---review and editing, H.O. and M.M.; visualization, A.S.; supervision, M.M.; project administration, M.M.; funding acquisition, M.M. All authors have read and agreed to the published version of the manuscript.}

\funding{This research was funded by the Carl Zeiss Foundation (Carl-Zeiss-Stiftung) as part of the OrthoKI project, grant number P2022-07-009.}

\institutionalreview{The study was conducted in accordance with the Declaration of Helsinki. The two measurement studies that provided the analyzed IMU recordings were approved by the Ethics Committee of Ulm University of Applied Sciences (protocol codes 2021-01 and 2024-01).}

\informedconsent{Informed consent was obtained from all subjects involved in the study.}

\dataavailability{A subset of the analyzed data is publicly available via Zenodo at \url{https://zenodo.org/records/15729056}. The remaining data are not publicly accessible but can be made available by the corresponding author upon reasonable request and subject to institutional data sharing agreements. All scripts and source code required to reproduce the methods are openly available at \url{https://github.com/ai-for-sensor-data-analytics-ulm/imu_augment_sim}.}

\conflictsofinterest{The authors declare no conflicts of interest. The funders had no role in the design of the study; in the collection, analyses, or interpretation of data; in the writing of the manuscript; or in the decision to publish the results.}

\acknowledgments{During the preparation of this manuscript, the authors used Claude (Anthropic) for language editing and to support the structuring of the text. The authors have reviewed and edited the output and take full responsibility for the content of this publication.}

%%%%%%%%%%%%%%%%%%%%%%%%%%%%%%%%%%%%%%%%%%
\isPreprints{}{% This command is only used for ``preprints''.
\begin{adjustwidth}{-\extralength}{0cm}
} % If the paper is ``preprints'', please uncomment this parenthesis.

\reftitle{References}

% Please provide either the correct journal abbreviation (e.g. according to the “The list of Title Word Abbreviations” https://portal.issn.org/ltwa) or the full name of the journal. 
% Citations and references in the Supplementary Materials are permitted provided that they also appear in the reference list here. 

\bibliography{sample.bib}

\begin{thebibliography}{999}

\bibitem[Ashari et~al.(2016)Ashari, Hamid, Hussain, and Hill]{ashari_effectiveness_2016}
Ashari, A.; Hamid, T.A.; Hussain, M.R.; Hill, K.D.
\newblock Effectiveness of {Individualized} {Home}-{Based} {Exercise} on {Turning} and {Balance} {Performance} {Among} {Adults} {Older} than 50 yrs.
\newblock {\em American Journal of Physical Medicine \& Rehabilitation} {\bf 2016}, {\em 95},~355--365.
\newblock {\url{https://doi.org/10.1097/phm.0000000000000388}}.

\bibitem[Latham et~al.(2014)Latham, Harris, Bean, Heeren, Goodyear, Zawacki, Heislein, Mustafa, Pardasaney, Giorgetti, Holt, Goehring, and Jette]{latham_effect_2014}
Latham, N.K.; Harris, B.A.; Bean, J.F.; Heeren, T.; Goodyear, C.; Zawacki, S.; Heislein, D.M.; Mustafa, J.; Pardasaney, P.; Giorgetti, M.;  et~al.
\newblock Effect of a {Home}-{Based} {Exercise} {Program} on {Functional} {Recovery} {Following} {Rehabilitation} {After} {Hip} {Fracture}: {A} {Randomized} {Clinical} {Trial}.
\newblock {\em JAMA} {\bf 2014}, {\em 311},~700--708.
\newblock {\url{https://doi.org/10.1001/jama.2014.469}}.

\bibitem[Gelaw et~al.(2020)Gelaw, Janakiraman, Gebremeskel, and Ravichandran]{gelaw_effectiveness_2020}
Gelaw, A.Y.; Janakiraman, B.; Gebremeskel, B.F.; Ravichandran, H.
\newblock Effectiveness of {Home}-based rehabilitation in improving physical function of persons with {Stroke} and other physical disability: {A} systematic review of randomized controlled trials.
\newblock {\em Journal of Stroke and Cerebrovascular Diseases: The Official Journal of National Stroke Association} {\bf 2020}, {\em 29},~104800.
\newblock {\url{https://doi.org/10.1016/j.jstrokecerebrovasdis.2020.104800}}.

\bibitem[Flynn et~al.(2019)Flynn, Allen, Dennis, Canning, and Preston]{flynn_home-based_2019}
Flynn, A.; Allen, N.E.; Dennis, S.; Canning, C.G.; Preston, E.
\newblock Home-based prescribed exercise improves balance-related activities in people with {Parkinson}'s disease and has benefits similar to centre-based exercise: a systematic review.
\newblock {\em Journal of Physiotherapy} {\bf 2019}, {\em 65},~189--199.
\newblock {\url{https://doi.org/10.1016/j.jphys.2019.08.003}}.

\bibitem[Argent et~al.(2018)Argent, Daly, and Caulfield]{argent_patient_2018}
Argent, R.; Daly, A.; Caulfield, B.
\newblock Patient {Involvement} {With} {Home}-{Based} {Exercise} {Programs}: {Can} {Connected} {Health} {Interventions} {Influence} {Adherence}?
\newblock {\em JMIR mHealth and uHealth} {\bf 2018}, {\em 6},~e47.
\newblock {\url{https://doi.org/10.2196/mhealth.8518}}.

\bibitem[Faber et~al.(2015)Faber, Andersen, Sevel, Thorborg, Bandholm, and Rathleff]{faber_majority_2015}
Faber, M.; Andersen, M.H.; Sevel, C.; Thorborg, K.; Bandholm, T.; Rathleff, M.
\newblock The majority are not performing home-exercises correctly two weeks after their initial instruction—an assessor-blinded study.
\newblock {\em PeerJ} {\bf 2015}, {\em 3},~e1102.
\newblock {\url{https://doi.org/10.7717/peerj.1102}}.

\bibitem[Lang et~al.(2022)Lang, McLelland, MacDonald, and Hamilton]{lang_digital_2022}
Lang, S.; McLelland, C.; MacDonald, D.; Hamilton, D.F.
\newblock Do digital interventions increase adherence to home exercise rehabilitation? {A} systematic review of randomised controlled trials.
\newblock {\em Archives of Physiotherapy} {\bf 2022}, {\em 12},~24.
\newblock {\url{https://doi.org/10.1186/s40945-022-00148-z}}.

\bibitem[Spilz and Munz(2023)]{spilz_automatic_2023}
Spilz, A.; Munz, M.
\newblock Automatic {Assessment} of {Functional} {Movement} {Screening} {Exercises} with {Deep} {Learning} {Architectures}.
\newblock {\em Sensors} {\bf 2023}, {\em 23},~5.
\newblock {\url{https://doi.org/10.3390/s23010005}}.

\bibitem[Cook et~al.(2014{\natexlab{a}})Cook, Burton, Hoogenboom, and Voight]{cook_functional_2014}
Cook, G.; Burton, L.; Hoogenboom, B.J.; Voight, M.
\newblock Functional movement screening: the use of fundamental movements as an assessment of function - part 1.
\newblock {\em International Journal of Sports Physical Therapy} {\bf 2014}, {\em 9},~396--409.

\bibitem[Cook et~al.(2014{\natexlab{b}})Cook, Burton, Hoogenboom, and Voight]{cook_functional_2014-1}
Cook, G.; Burton, L.; Hoogenboom, B.J.; Voight, M.
\newblock Functional movement screening: the use of fundamental movements as an assessment of function-part 2.
\newblock {\em International Journal of Sports Physical Therapy} {\bf 2014}, {\em 9},~549--563.

\bibitem[Xing et~al.(2022)Xing, Shen, Cao, Zong, Zhao, and Shen]{xing_functional_2022}
Xing, Q.J.; Shen, Y.Y.; Cao, R.; Zong, S.X.; Zhao, S.X.; Shen, Y.F.
\newblock Functional movement screen dataset collected with two {Azure} {Kinect} depth sensors.
\newblock {\em Scientific Data} {\bf 2022}, {\em 9},~104.
\newblock {\url{https://doi.org/10.1038/s41597-022-01188-7}}.

\bibitem[Wu et~al.(2020)Wu, Lee, Hsu, Ho, and Liang]{wu_development_2020}
Wu, W.L.; Lee, M.H.; Hsu, H.T.; Ho, W.H.; Liang, J.M.
\newblock Development of an {Automatic} {Functional} {Movement} {Screening} {System} with {Inertial} {Measurement} {Unit} {Sensors}.
\newblock {\em Applied Sciences} {\bf 2020}, {\em 11},~96.
\newblock {\url{https://doi.org/10.3390/app11010096}}.

\bibitem[Scheurer et~al.(2020)Scheurer, Tedesco, O’Flynn, and Brown]{scheurer_comparing_2020}
Scheurer, S.; Tedesco, S.; O’Flynn, B.; Brown, K.N.
\newblock Comparing {Person}-{Specific} and {Independent} {Models} on {Subject}-{Dependent} and {Independent} {Human} {Activity} {Recognition} {Performance}.
\newblock {\em Sensors} {\bf 2020}, {\em 20},~3647.
\newblock {\url{https://doi.org/10.3390/s20133647}}.

\bibitem[Kianifar et~al.(2017)Kianifar, Lee, Raina, and Kulic]{kianifar_automated_2017}
Kianifar, R.; Lee, A.; Raina, S.; Kulic, D.
\newblock Automated {Assessment} of {Dynamic} {Knee} {Valgus} and {Risk} of {Knee} {Injury} {During} the {Single} {Leg} {Squat}.
\newblock {\em IEEE Journal of Translational Engineering in Health and Medicine} {\bf 2017}, {\em 5},~2100213.
\newblock {\url{https://doi.org/10.1109/JTEHM.2017.2736559}}.

\bibitem[Sharifi~Renani et~al.(2021)Sharifi~Renani, Eustace, Myers, and Clary]{sharifi_renani_use_2021}
Sharifi~Renani, M.; Eustace, A.M.; Myers, C.A.; Clary, C.W.
\newblock The {Use} of {Synthetic} {IMU} {Signals} in the {Training} of {Deep} {Learning} {Models} {Significantly} {Improves} the {Accuracy} of {Joint} {Kinematic} {Predictions}.
\newblock {\em Sensors} {\bf 2021}, {\em 21},~5876.
\newblock {\url{https://doi.org/10.3390/s21175876}}.

\bibitem[Mundt et~al.(2020)Mundt, Koeppe, David, Witter, Bamer, Potthast, and Markert]{mundt_estimation_2020}
Mundt, M.; Koeppe, A.; David, S.; Witter, T.; Bamer, F.; Potthast, W.; Markert, B.
\newblock Estimation of {Gait} {Mechanics} {Based} on {Simulated} and {Measured} {IMU} {Data} {Using} an {Artificial} {Neural} {Network}.
\newblock {\em Frontiers in Bioengineering and Biotechnology} {\bf 2020}, {\em 8},~41.
\newblock {\url{https://doi.org/10.3389/fbioe.2020.00041}}.

\bibitem[Uhlenberg et~al.(2024)Uhlenberg, Ole~Haeusler, and Amft]{uhlenberg_synhar_2024}
Uhlenberg, L.; Ole~Haeusler, L.; Amft, O.
\newblock {SynHAR}: {Augmenting} {Human} {Activity} {Recognition} {With} {Synthetic} {Inertial} {Sensor} {Data} {Generated} {From} {Human} {Surface} {Models}.
\newblock {\em IEEE Access} {\bf 2024}, {\em 12},~194839--194858.
\newblock {\url{https://doi.org/10.1109/ACCESS.2024.3513477}}.

\bibitem[Kwon et~al.(2020)Kwon, Tong, Haresamudram, Gao, Abowd, Lane, and Plötz]{kwon_imutube_2020}
Kwon, H.; Tong, C.; Haresamudram, H.; Gao, Y.; Abowd, G.D.; Lane, N.D.; Plötz, T.
\newblock {IMUTube}: {Automatic} {Extraction} of {Virtual} on-body {Accelerometry} from {Video} for {Human} {Activity} {Recognition}.
\newblock {\em Proc. ACM Interact. Mob. Wearable Ubiquitous Technol.} {\bf 2020}, {\em 4},~87:1--87:29.
\newblock {\url{https://doi.org/10.1145/3411841}}.

\bibitem[Zolfaghari et~al.(2024)Zolfaghari, Rey, Ray, Kim, Suh, and Lukowicz]{zolfaghari_sensor_2024}
Zolfaghari, P.; Rey, V.F.; Ray, L.; Kim, H.; Suh, S.; Lukowicz, P.
\newblock Sensor {Data} {Augmentation} from {Skeleton} {Pose} {Sequences} for {Improving} {Human} {Activity} {Recognition}.
\newblock In Proceedings of the 2024 {International} {Conference} on {Activity} and {Behavior} {Computing} ({ABC}),  2024, pp. 1--8.
\newblock {\url{https://doi.org/10.1109/ABC61795.2024.10652200}}.

\bibitem[Norgaard et~al.(2018)Norgaard, Saeedi, Sasani, and Gebremedhin]{norgaard_synthetic_2018}
Norgaard, S.; Saeedi, R.; Sasani, K.; Gebremedhin, A.H.
\newblock Synthetic {Sensor} {Data} {Generation} for {Health} {Applications}: {A} {Supervised} {Deep} {Learning} {Approach}.
\newblock In Proceedings of the 2018 40th {Annual} {International} {Conference} of the {IEEE} {Engineering} in {Medicine} and {Biology} {Society} ({EMBC}),  2018, pp. 1164--1167.
\newblock {\url{https://doi.org/10.1109/EMBC.2018.8512470}}.

\bibitem[Mohammadzadeh et~al.(2025)Mohammadzadeh, Ghadami, Taheri, and Behzadipour]{mohammadzadeh_cgan-based_2025}
Mohammadzadeh, M.; Ghadami, A.; Taheri, A.; Behzadipour, S.
\newblock {cGAN}-based high dimensional {IMU} sensor data generation for enhanced human activity recognition in therapeutic activities.
\newblock {\em Biomedical Signal Processing and Control} {\bf 2025}, {\em 103},~107476.
\newblock {\url{https://doi.org/10.1016/j.bspc.2024.107476}}.

\bibitem[Zhao et~al.(2022)Zhao, Obonyo, and Yin]{zhao_improving_2022}
Zhao, J.; Obonyo, E.; Yin, Q.
\newblock Improving posture recognition among construction workers through data augmentation with generative adversarial network.
\newblock {\em IOP Conference Series: Earth and Environmental Science} {\bf 2022}, {\em 1101},~092005.
\newblock {\url{https://doi.org/10.1088/1755-1315/1101/9/092005}}.

\bibitem[Leng et~al.(2023)Leng, Kwon, and Plötz]{leng_generating_2023}
Leng, Z.; Kwon, H.; Plötz, T.
\newblock Generating {Virtual} {On}-body {Accelerometer} {Data} from {Virtual} {Textual} {Descriptions} for {Human} {Activity} {Recognition}.
\newblock In Proceedings of the Proceedings of the 2023 {ACM} {International} {Symposium} on {Wearable} {Computers}, New York, NY, USA,  2023; {ISWC} '23, pp. 39--43.
\newblock {\url{https://doi.org/10.1145/3594738.3611361}}.

\bibitem[Leng et~al.(2024)Leng, Bhattacharjee, Rajasekhar, Zhang, Bruda, Kwon, and Plötz]{leng_imugpt_2024}
Leng, Z.; Bhattacharjee, A.; Rajasekhar, H.; Zhang, L.; Bruda, E.; Kwon, H.; Plötz, T.
\newblock {IMUGPT} 2.0: {Language}-{Based} {Cross} {Modality} {Transfer} for {Sensor}-{Based} {Human} {Activity} {Recognition}.
\newblock {\em Proceedings of the ACM on Interactive, Mobile, Wearable and Ubiquitous Technologies} {\bf 2024}, {\em 8},~1--32.
\newblock {\url{https://doi.org/10.1145/3678545}}.

\bibitem[Chandrasekaran et~al.(2023)Chandrasekaran, Francik, and Makris]{chandrasekaran_gait_2023}
Chandrasekaran, M.; Francik, J.; Makris, D.
\newblock Gait {Data} {Augmentation} using {Physics}-{Based} {Biomechanical} {Simulation},  2023.
\newblock arXiv:2307.08092 [cs], {\url{https://doi.org/10.48550/arXiv.2307.08092}}.

\bibitem[Tang et~al.(2024)Tang, He, Xu, Tan, Wang, Zhou, and Jiang]{tang_synthetic_2024}
Tang, J.; He, B.; Xu, J.; Tan, T.; Wang, Z.; Zhou, Y.; Jiang, S.
\newblock Synthetic {IMU} {Datasets} and {Protocols} {Can} {Simplify} {Fall} {Detection} {Experiments} and {Optimize} {Sensor} {Configuration}.
\newblock {\em IEEE Transactions on Neural Systems and Rehabilitation Engineering} {\bf 2024}, {\em 32},~1233--1245.
\newblock {\url{https://doi.org/10.1109/TNSRE.2024.3370396}}.

\bibitem[Oishi et~al.(2026)Oishi, Birch, Roggen, and Lago]{oishi_physically_2026}
Oishi, N.; Birch, P.; Roggen, D.; Lago, P.
\newblock Physically {Plausible} {Data} {Augmentations} for {Wearable} {IMU}-{Based} {Human} {Activity} {Recognition} {Using} {Physics} {Simulation}.
\newblock {\em IEEE Sensors Journal} {\bf 2026}, {\em 26},~10445--10457.
\newblock {\url{https://doi.org/10.1109/JSEN.2026.3661047}}.

\bibitem[Dorschky et~al.(2020)Dorschky, Nitschke, Martindale, van~den Bogert, Koelewijn, and Eskofier]{dorschky_cnn-based_2020}
Dorschky, E.; Nitschke, M.; Martindale, C.F.; van~den Bogert, A.J.; Koelewijn, A.D.; Eskofier, B.M.
\newblock {CNN}-{Based} {Estimation} of {Sagittal} {Plane} {Walking} and {Running} {Biomechanics} {From} {Measured} and {Simulated} {Inertial} {Sensor} {Data}.
\newblock {\em Frontiers in Bioengineering and Biotechnology} {\bf 2020}, {\em 8},~604.
\newblock {\url{https://doi.org/10.3389/fbioe.2020.00604}}.

\bibitem[Spilz et~al.(2025)Spilz, Oppel, Werner, Stucke-Straub, Capanni, and Munz]{spilz_gaitex_2025}
Spilz, A.; Oppel, H.; Werner, J.; Stucke-Straub, K.; Capanni, F.; Munz, M.
\newblock {GAITEX}: {Human} motion dataset of impaired gait and rehabilitation exercises using inertial and optical sensors.
\newblock {\em Scientific Data} {\bf 2025}, {\em 13},~11.
\newblock {\url{https://doi.org/10.1038/s41597-025-06439-x}}.

\bibitem[Shultz et~al.(2013)Shultz, Anderson, Matheson, Marcello, and Besier]{shultz_test-retest_2013}
Shultz, R.; Anderson, S.C.; Matheson, G.O.; Marcello, B.; Besier, T.
\newblock Test-{Retest} and {Interrater} {Reliability} of the {Functional} {Movement} {Screen}.
\newblock {\em Journal of Athletic Training} {\bf 2013}, {\em 48},~331--336.
\newblock {\url{https://doi.org/10.4085/1062-6050-48.2.11}}.

\bibitem[Roetenberg et~al.(2009)Roetenberg, Luinge, and Slycke]{roetenberg}
Roetenberg, D.; Luinge, H.; Slycke, P.
\newblock Xsens MVN: Full 6DOF human motion tracking using miniature inertial sensors.
\newblock {\em Xsens Motion Technol. BV Tech. Rep.} {\bf 2009}, {\em 3}.

\bibitem[Madgwick et~al.(2011)Madgwick, Harrison, and Vaidyanathan]{madgwick_estimation_2011}
Madgwick, S.O.H.; Harrison, A.J.L.; Vaidyanathan, R.
\newblock Estimation of {IMU} and {MARG} orientation using a gradient descent algorithm.
\newblock In Proceedings of the 2011 {IEEE} {International} {Conference} on {Rehabilitation} {Robotics},  2011, pp. 1--7.
\newblock {\url{https://doi.org/10.1109/ICORR.2011.5975346}}.

\bibitem[Paulich et~al.(2018)Paulich, Schepers, Rudigkeit, and Bellusci]{paulich_xsens_2018}
Paulich, M.; Schepers, M.; Rudigkeit, N.; Bellusci, G.
\newblock Xsens {MTw} {Awinda}: {Miniature} {Wireless} {Inertial}-{Magnetic} {Motion} {Tracker} for {Highly} {Accurate} {3D} {Kinematic} {Applications}.
\newblock Technical report, XSENS Technologies B.V.,  2018.

\bibitem[Delp et~al.(2007)Delp, Anderson, Arnold, Loan, Habib, John, Guendelman, and Thelen]{delp_opensim_2007}
Delp, S.L.; Anderson, F.C.; Arnold, A.S.; Loan, P.; Habib, A.; John, C.T.; Guendelman, E.; Thelen, D.G.
\newblock {OpenSim}: open-source software to create and analyze dynamic simulations of movement.
\newblock {\em IEEE Transactions on Biomedical Engineering} {\bf 2007}, {\em 54},~1940--1950.
\newblock {\url{https://doi.org/10.1109/TBME.2007.901024}}.

\bibitem[Rajagopal et~al.(2016)Rajagopal, Dembia, DeMers, Delp, Hicks, and Delp]{rajagopal_full-body_2016}
Rajagopal, A.; Dembia, C.L.; DeMers, M.S.; Delp, D.D.; Hicks, J.L.; Delp, S.L.
\newblock Full-{Body} {Musculoskeletal} {Model} for {Muscle}-{Driven} {Simulation} of {Human} {Gait}.
\newblock {\em IEEE Transactions on Biomedical Engineering} {\bf 2016}, {\em 63},~2068--2079.
\newblock {\url{https://doi.org/10.1109/TBME.2016.2586891}}.

\bibitem[Van~der Maaten and Hinton(2008)]{tsne}
Van~der Maaten, L.; Hinton, G.
\newblock Visualizing data using t-SNE.
\newblock {\em Journal of Machine Learning Research} {\bf 2008}, {\em 9}.

\bibitem[Shoemake(1985)]{shoemake_animating_1985}
Shoemake, K.
\newblock Animating rotation with quaternion curves.
\newblock {\em SIGGRAPH Comput. Graph.} {\bf 1985}, {\em 19},~245--254.
\newblock {\url{https://doi.org/10.1145/325165.325242}}.

\bibitem[Nitschke et~al.(2024)Nitschke, Dorschky, Leyendecker, Eskofier, and Koelewijn]{nitschke_estimating_2024}
Nitschke, M.; Dorschky, E.; Leyendecker, S.; Eskofier, B.M.; Koelewijn, A.D.
\newblock Estimating {3D} kinematics and kinetics from virtual inertial sensor data through musculoskeletal movement simulations.
\newblock {\em Frontiers in Bioengineering and Biotechnology} {\bf 2024}, {\em 12},~1285845.
\newblock {\url{https://doi.org/10.3389/fbioe.2024.1285845}}.

\bibitem[Gao et~al.(2017)Gao, Xing, Xie, Wu, and Geng]{gaoDeepLabelDistribution2017a}
Gao, B.B.; Xing, C.; Xie, C.W.; Wu, J.; Geng, X.
\newblock Deep {Label} {Distribution} {Learning} {With} {Label} {Ambiguity}.
\newblock {\em IEEE Transactions on Image Processing} {\bf 2017}, {\em 26},~2825--2838.
\newblock {\url{https://doi.org/10.1109/TIP.2017.2689998}}.

\end{thebibliography}
===

% If authors have biography, please use the format below
%\section{Short Biography of Authors}
%\bio
%{\raisebox{-0.35cm}{\includegraphics[width=3.5cm,height=5.3cm,clip,keepaspectratio]{Definitions/author1.pdf}}}
%{\textbf{Firstname Lastname} Biography of first author}
%
%\bio
%{\raisebox{-0.35cm}{\includegraphics[width=3.5cm,height=5.3cm,clip,keepaspectratio]{Definitions/author2.jpg}}}
%{\textbf{Firstname Lastname} Biography of second author}

% For the MDPI journals use author-date citation, please follow the formatting guidelines on http://www.mdpi.com/authors/references
% To cite two works by the same author:~\citepauthor{ref-journal-1a} (\citepyear{ref-journal-1a},~\citepyear{ref-journal-1b}). This produces: Whittaker (1967, 1975)
% To cite two works by the same author with specific pages:~\citepauthor{ref-journal-3a} (\citepyear{ref-journal-3a}, p. 328;~\citepyear{ref-journal-3b}, p.475). This produces: Wong (1999, p. 328; 2000, p. 475)

\PublishersNote{}
\isPreprints{}{% This command is only used for ``preprints''.
\end{adjustwidth}
} % If the paper is ``preprints'', please uncomment this parenthesis.
\end{document}